\documentclass[preprint,12pt]{elsarticle}
\usepackage{enumitem}
\usepackage{algorithm}
\usepackage{algpseudocode}
\usepackage{amsmath}
\usepackage{subcaption}
\usepackage{booktabs}
\usepackage{graphicx}
\usepackage{subcaption}
\usepackage{float}

\usepackage{amssymb}
\usepackage{amsmath}

\journal{}  

\makeatletter
\def\ps@pprintTitle{%
 \let\@oddhead\@empty
 \let\@evenhead\@empty
 \def\@oddfoot{\reset@font\hfil\thepage\hfil}
 \let\@evenfoot\@oddfoot}
\makeatother
\begin{document}

\begin{frontmatter}



\title{A High-Dimensional Feature Selection Algorithm Based on Multiobjective Differential Evolution}

\author[aff1]{Zhenxing Zhang\corref{cor1}}
\ead{ludongzx@ldu.edu.cn}
\cortext[cor1]{Corresponding author.}

\author[aff2]{Qianxiang An}
\author[aff2]{Yilei Wang}
\author[aff1]{Chenfeng Wu}
\author[aff2]{Baoling Dong}
\author[aff1]{Chunjie Zhou}

\affiliation[aff1]{
  organization={Department of Information and Electrical Engineering, Ludong University},
  city={Yantai},
  postcode={264025},
  state={Shandong},
  country={China}
}

\affiliation[aff2]{
  organization={Department of Computer Science, Qufu Normal University},
  city={Rizhao},
  postcode={276826},
  state={Shandong},
  country={China}
}

\begin{abstract}
Multiobjective feature selection seeks to determine the most discriminative feature subset by simultaneously optimizing two conflicting objectives: minimizing the number of selected features and the classification error rate. The goal is to enhance the model’s predictive performance and computational efficiency. However, feature redundancy and interdependence in high-dimensional data present considerable obstacles to the search efficiency of optimization algorithms and the quality of the resulting solutions. To tackle these issues, we propose a high-dimensional feature selection algorithm based on multiobjective differential evolution. First, a population initialization strategy is designed by integrating feature weights and redundancy indices, where the population is divided into four subpopulations to improve the diversity and uniformity of the initial population. Then, a multiobjective selection mechanism is developed, in which feature weights guide the mutation process. The solution quality is further enhanced through nondominated sorting, with preference given to solutions with lower classification error, effectively balancing global exploration and local exploitation. Finally, an adaptive grid mechanism is applied in the objective space to identify densely populated regions and detect duplicated solutions. Experimental results on $11$ UCI datasets of varying difficulty demonstrate that the proposed method significantly outperforms several state-of-the-art multiobjective feature selection approaches regarding feature selection performance.
\end{abstract}


\end{frontmatter}



\section{Introduction}
\label{sec1}
With the rapid development of the Internet, massive volumes of data have been generated through information exchange across both industrial and consumer applications~\cite{xue2023_1}. However, such datasets often exhibit numerous irrelevant and redundant features. Irrelevant features have negligible impact on the output, while redundant features are typically linear combinations of others, providing no additional discriminatory information. The presence of these features not only significantly increases the computational complexity of learning algorithms but may also negatively affect model performance~\cite{hashemi2021_2}. Feature selection (FS) has become a crucial technique in data mining and machine learning for effectively mitigating these issues. The objectives of feature selection often involve two conflicting aspects: on the one hand, minimizing the number of selected features to reduce computational cost; on the other hand, maximizing classification accuracy to ensure model performance. Consequently, FS is typically formulated as a Multiobjective Feature Selection (MOFS) problem, which aims to simultaneously reduce the number of features and improve classification performance~\cite{jiao2024_3}.\par
Multiobjective evolutionary algorithms (MOEAs) simulate the process of natural evolution to generate a set of nondominated solutions in a single run, thereby enabling efficient optimization in MOFS tasks~\cite{hashemi2021_2}. In contrast to conventional approaches, MOEAs provide substantial benefits in search capability, solution diversity, and global optimality~\cite{jiao2023_4,wang2023_5}. In general, based on different selection mechanisms, MOEAs can be broadly categorized into three types: dominance-based algorithms (e.g., NSGA-II), decomposition-based algorithms (e.g., MOEA/D), and indicator-based algorithms (e.g., IBEA)~\cite{pan2022_6,deb2002_7,zhang2007_8,tian2017_9}. Although these algorithms have their respective strengths, they still face several limitations in practical applications. Dominance-based algorithms, particularly those utilizing Pareto dominance, often focus on the central region of the Pareto front and tend to struggle with identifying feature subsets that offer high classification performance or involve a small number of selected features. Moreover, due to frequent occurrences of duplicated solutions, their diversity maintenance mechanisms are often ineffective for MOFS problems. A major challenge in decomposition-based algorithms is that the distribution of the obtained Pareto-optimal solutions heavily depends on the setting of the direction vectors. Specifying proper direction vectors is exceptionally demanding for MOFS, given the unknown geometry of the Pareto front~\cite{jiao2023_4}. On the other hand, although indicator-based algorithms can directly optimize both the distribution and convergence of solutions, they often incur high computational costs when applied to large-scale problems.\par
To address the above issues, scholars have performed comprehensive studies on MOEAs and have widely applied them to feature selection. Commonly used evolutionary algorithms for feature selection include Genetic Algorithm (GA)~\cite{deb1995_10}, Firefly Algorithm (FA)~\cite{fister2015_11}, Particle Swarm Optimization (PSO)~\cite{qin2020_12}, and Differential Evolution (DE)~\cite{kukkonen2005_13}. Among these, DE has exhibited exceptional performance in feature selection tasks due to its simple structure, intuitive parameter settings, and high search efficiency, and has become a significant research direction in this field~\cite{wang2023_5}. For instance, Pan et al.~\cite{pan2022_6} proposed a multiobjective differential evolution algorithm based on a competitive mechanism, which demonstrated robust efficacy in optimizing feature subset selection. Wang et al.~\cite{wang2023_14} developed a DE-based method that incorporates feature clustering to explore multiple optimal subsets with similar classification performance. Hu et al.~\cite{hu2023_15} introduced two network-structured DE algorithms, 
which construct feature networks to model inter-feature relationships and enhance the search process. Agrawal et al.~\cite{agrawal2023_16} designed a multimodal DE algorithm that employs probabilistic initialization and nearest-neighbor clustering to explore the feature space, and utilizes a Stagnation Convergence Archive to retain high-quality solutions. Yu et al.~\cite{yu2025_17} proposed a method that combines mutual information-based ranking with multipopulation DE, achieving excellent performance on multiple datasets, especially in high-dimensional scenarios. Yu et al.~\cite{yu2024_18} presented a reinforcement learning-based multiobjective differential evolution algorithm, where mutation operations are dynamically adjusted through a learning framework, and a Pareto front relearning strategy is employed to preserve solution diversity. Despite the effectiveness of these methods, several key issues persist unresolved. First, the randomness in population initialization often leads to imbalanced individual distribution, affecting search efficiency. Second, traditional DE algorithms do not sufficiently consider inter-feature correlations during the mutation process, which limits their search capability. Lastly, during the optimization process, feature subset duplication and dense solution regions may lead to a decline in population diversity, thereby restricting the exploration of the search space and causing premature convergence.\par
In this paper, we propose a novel feature selection algorithm based on differential evolution, which is composed of three core modules. The first module implements an initialization strategy that enhances population diversity and improves the uniformity of initial solution distribution. This is achieved by dividing the population into four subpopulations based on feature importance and redundancy. The second module introduces an individual update mechanism, which adjusts the feature correlation weight matrix and integrates a redundancy index. This mechanism filters uninformative features and enhances both diversity and global search ability. The third module applies an adaptive grid strategy in the objective space to dynamically assess and refine grid density, thereby improving the distribution uniformity of solutions and expanding search space coverage. To assess the efficacy of the proposed algorithm, experiments were conducted on 11 UCI datasets. The results demonstrate that the proposed method outperforms existing feature selection approaches across multiple datasets.\par
The main contributions of this work are summarized as follows:
\begin{itemize}[itemsep=0pt, parsep=0pt, topsep=0pt, partopsep=0pt, left=1.5em]
\item We introduce a Fuzzy Cognitive Map (FCM)-based approach to compute correlation weights between features and the target label, enabling a more informed assessment of feature importance. FCM models the influence relationships of features on the target label and generates a feature correlation weight matrix $Q$, which is used to evaluate the importance of each feature in the classification task. This approach proficiently captures the contribution of different features to target prediction, providing a more rational evaluation basis for feature selection.
\item A Weighted-Redundancy Balanced Initialization (WRBI) strategy is designed. The population is divided into four subpopulations: a high-quality subpopulation containing high-weight features, a low-quality subpopulation that ignores redundancy, an elite subpopulation with redundancy removed, and an elite subpopulation without redundancy removal. This strategy enhances the diversity and distribution uniformity of the initial population, establishing a robust basis for subsequent optimization.
\item A Mutation-Selection Based Individual Update (MSBIU) mechanism is proposed to guide individual mutation toward more promising regions of the search space. This mechanism adjusts the feature correlation weight matrix $Q$ via mean shift, retaining only those features that make positive contributions to the optimization objectives. In addition, it incorporates the feature redundancy index $A$ to guide feature filtering, effectively balancing global exploration and local exploitation. This strategy significantly enhances both the diversity and quality of solutions, improves the rationality and efficiency of feature selection, and substantially enhances the overall optimization performance of the population.
\item A Feature-Optimized Adaptive Grid Mechanism (FOAGM) is developed to regulate the distribution of solutions in the objective space. FOAGM dynamically evaluates solution density and applies refinement strategies in dense grid regions: prioritizing nondominated solutions, selecting those with the lowest classification error if necessary, and refining duplicates by replacing redundant features with low importance with more informative alternatives. This approach alleviates solution concentration, improves search coverage, and enhances global and local search capabilities in multiobjective optimization.
\end{itemize} \par
Based on the above contributions, the structure of the paper is arranged as follows. Section~\ref{sec:BGRW} introduces the background and related work. Section~\ref{sec:Proposed} presents the proposed algorithm. Section~\ref{sec:Experiments} details the experimental setup, followed by Section ~\ref{sec:Results}, which discusses the experimental results and verifies the effectiveness of the method. Finally, Section~\ref{sec:conclusion} concludes the paper with a summary of the principal conclusions.



\section{Background and Related Work}\label{sec:BGRW}
\label{sec1}
This section begins by defining multiobjective optimization problems (MOP) and introducing related concepts. It then describes the classical differential evolution algorithm along with commonly used mutation operators. Subsequently, it introduces the fundamentals and modeling capabilities of Fuzzy Cognitive Map. Finally, it systematically reviews existing research on multiobjective feature selection algorithms.


\subsection{Multiobjective Optimization for Feature Selection}
\label{subsec1}

In multiobjective optimization, two or more conflicting objectives are generally optimized simultaneously~\cite{yue2018_19}. Improvements in one objective may adversely affect others, making it necessary to seek a set of trade-off solutions—known as the Pareto optimal set. A general minimization-based multiobjective optimization problem can be formulated as shown in Eq.(\ref{eq1}).
\begin{equation}
\begin{aligned}
& \min \vec{f}(\vec{x}) = \left( f_1(\vec{x}), f_2(\vec{x}), \ldots, f_L(\vec{x}) \right) \\
& \text{s.t. } g_i(\vec{x}) \leq 0, \quad i=1,2,\ldots,g
\end{aligned}
\label{eq1}
\end{equation}
where $\vec{x}$ denotes a candidate solution in the solution space $\Omega$, $D$ is the dimensionality of the solution, $L$ represents the number of objective functions to be optimized, and $g_i(\vec{x})$ denotes the corresponding constraints. In the context of feature selection, the two primary objectives are to minimize the number of selected features and the classification error rate simultaneously. In evolutionary computation-based methods, a feature selection solution is typically represented as a $D$-dimensional vector, as shown in Eq.(\ref{eq2}).
\begin{equation}
\vec{x} = (x_1, x_2, \ldots, x_D), \quad x_j \in \{0,1\}, \quad j = 1,2,\ldots,D
\label{eq2}
\end{equation} \par
Each component $x_j$ of the vector $\vec{x}$ corresponds to a specific feature. When $x_j = 1$, the $j$-th feature is selected; when $x_j = 0$, it is discarded. By counting the number of elements in the vector with a value of 1 (denoted as $m$), the proportion of selected features to the total number of features, referred to as the Feature Ratio (FR), can be obtained. The computation is shown in Eq.(\ref{eq3}).
\begin{equation}
\text{FR} = \frac{m}{D}
\label{eq3}
\end{equation} \par
In addition, to evaluate classification performance, the error rate (ER), defined as the proportion of incorrectly classified samples to the total number of samples. The computation is shown in Eq.(\ref{eq4}).
\begin{equation}
\text{ER} = \frac{\text{FP} + \text{FN}}{\text{TP} + \text{TN} + \text{FP} + \text{FN}}
\label{eq4}
\end{equation}
where TP and TN represent true positives and true negatives, respectively, and FP and FN denote false positives and false negatives, respectively. The feature selection problem can be formulated as a multiobjective optimization problem, with the goal of simultaneously minimizing the feature ratio and the classification error rate, as shown in Eq.(\ref{eq5}).
\begin{equation}
\begin{aligned}
& \min F(\vec{x}) = \left( f_{\text{FR}}(\vec{x}), f_{\text{ER}}(\vec{x}) \right) \\
& \text{where}~\vec{x} = (x_1, \ldots, x_D) \in \Omega = \{0,1\}^D
\end{aligned}
\label{eq5}
\end{equation}\par
As a result, the feature selection problem can be explicitly formulated as an MOP, where both FR and ER are constrained within (0, 1), facilitating more effective optimization.
\subsection{Differential Evolution}
\label{subsec1}
Differential Evolution (DE) is a population-based evolutionary algorithm proposed by Price and Storn~\cite{kukkonen2005_13}. DE, inspired by the process of natural evolution, optimizes the objective function by iteratively evolving a population of candidate solutions through mutation, crossover, and selection operations. Initially, a population of $NP$ individuals is randomly generated in DE. Each individual $X_{i,G}$ is represented as a $D$-dimensional vector, encoding a candidate solution to the optimization problem. The mathematical representation is shown in Eq.(\ref{eq6}).
\begin{equation}
\begin{aligned}
X_{i,G} &= (x_{i,1,G}, x_{i,2,G}, \ldots, x_{i,D,G}), \quad i = 1,2,\ldots,NP \\
x_{i,j,0} &= x_{\min,j} + \text{rand}(0,1) \cdot (x_{\max,j} - x_{\min,j}), \quad j = 1,2,\ldots,D
\end{aligned}
\label{eq6}
\end{equation}
where $G$ denotes the current generation, $i$ is the index of the individual, and $j$ is the index of the variable. $x_{\max,j}$ and $x_{\min,j}$ represent the upper and lower bounds of the $j$-th variable, respectively. The function $\text{rand}(0,1)$ generates a random number between 0 and 1, ensuring uniform distribution of the variables within the search space.\par
After initialization, DE enters the evolutionary phase, beginning with the mutation operation, which generates a mutant vector $V_{i,G}$ that determines the search direction. Different mutation strategies have a profound influence on the search capability of the algorithm. Five commonly used mutation strategies are defined in Eqs.(\ref{eq7})--(\ref{eq11}).
\setcounter{equation}{6}

\begin{flushleft}

DE/rand/1: 
\begin{equation}
V_{i,G} = X_{r1,G} + F \cdot (X_{r2,G} - X_{r3,G})
\label{eq7}
\end{equation}

DE/rand/2: 
\begin{equation}
V_{i,G} = X_{r1,G} + F \cdot (X_{r2,G} - X_{r3,G}) + F \cdot (X_{r4,G} - X_{r5,G})
\label{eq8}
\end{equation}

DE/best/1: 
\begin{equation}
V_{i,G} = X_{\text{best},G} + F \cdot (X_{r1,G} - X_{r2,G})
\label{eq9}
\end{equation}

DE/best/2: 
\begin{equation}
V_{i,G} = X_{\text{best},G} + F \cdot (X_{r1,G} - X_{r2,G}) + F \cdot (X_{r3,G} - X_{r4,G})
\label{eq10}
\end{equation}

DE/current-to-best/1:
\begin{equation}
\begin{aligned}
V_{i,G} &= X_{i,G} + F \cdot (X_{\text{best},G} - X_{i,G}) + F \cdot (X_{r1,G} - X_{r2,G})
\end{aligned}
\label{eq11}
\end{equation}

\end{flushleft}
where $X_{\text{best},G}$ represents the best individual in the current generation, while $X_{1,G}$, $X_{2,G}$, $X_{3,G}$, $X_{4,G}$, and $X_{5,G}$ are randomly selected, distinct individuals from the population. $F$ is the scaling factor that controls the mutation strength and is commonly set within the range of $(0,1)$.

The mutated individual is then combined with the original individual through a crossover operator to generate a trial vector $U_{i,G}$. The crossover operation enhances population diversity and is performed according to Eq.(\ref{eq12}).
\begin{equation}
u_{i,j,G} =
\begin{cases}
v_{i,j,G}, & \text{if } \text{rand}(0,1) \leq Cr \text{ or } j = j_{\text{rand}} \\
x_{i,j,G}, & \text{otherwise}
\end{cases}
\label{eq12}
\end{equation}
where $Cr$ is the crossover probability, which controls the extent to which the mutant vector is incorporated into the trial vector. Its value is usually set between 0 and 1. The variable $j_{\text{rand}}$ ensures that at least one dimension undergoes mutation, thereby preventing the individual from remaining identical to its parent. After crossover, DE applies a selection operator to determine which individuals are allowed to proceed to the next generation. Selection is based on a fitness comparison strategy, retaining the individual with superior fitness. The selection process is defined in Eq.(\ref{eq13}).
\begin{equation}
X_{i,G+1} =
\begin{cases}
U_{i,G}, & \text{if } f(U_{i,G}) \leq f(X_{i,G}) \\
X_{i,G}, & \text{otherwise}
\end{cases}
\label{eq13}
\end{equation}\par
The objective function is denoted as $f(\cdot)$, where smaller values indicate better solutions in minimization problems. The selection operation promotes the gradual convergence of the population toward better solutions during the evolutionary process. The DE algorithm iteratively executes mutation, crossover, and selection operations until a termination condition is satisfied. Common termination criteria include: First, reaching a maximum number of generations $G_{\text{max}}$; Second, the fitness of the best individual remaining unchanged over several consecutive generations; and Third, the objective function value reaching a predefined optimization threshold.
\subsection{Fuzzy Cognitive Map}
Fuzzy Cognitive Map (FCM) is an advanced fuzzy modeling technique that integrates the core principles of fuzzy logic with the adaptive learning capabilities of neural networks, making it highly suitable for representing and analyzing complex systems~\cite{kosko1986_20}. Structurally, an FCM consists of a set of nodes (also referred to as concepts) and directed, weighted connections that signify the causal influence among these concepts. Each node encapsulates a specific concept or variable from the real world, while the weighted edges characterize the strength and direction of interdependencies between these concepts. Owing to its computational efficiency in numerical inference, transparent interpretability, and capacity for intuitive knowledge encoding, the FCM has been successfully employed in a broad spectrum of domains. These include, but are not limited to, medical disorder analysis and classification, strategic decision support systems, as well as tasks related to time series analysis and forecasting.
\subsection{Existing Multiobjective Feature Selection Algorithms}
\label{subsec1}
Under the framework of Evolutionary Multiobjective Optimization (EMO), numerous evolutionary optimization approaches have been proposed in recent years to address the feature selection problem. To better understand the current progress in this field, this section reviews and analyzes several representative multiobjective feature selection approaches. Xue et al.~\cite{xue2014_21} pioneered a feature selection method using multiobjective differential evolution (DEMOFS). Experimental results on nine datasets demonstrated that DEMOFS outperformed linear forward selection and greedy stepwise backward selection methods. To accelerate convergence, Zhang et al.~\cite{zhang2015_22} proposed using the nondominated vector among three randomly selected vectors as the base vector $x_{base,i}$ for individual $x_i$, in order to guide the search direction of differential evolution. Salesi et al.~\cite{salesi2021_23} proposed a two-stage feature selection approach based on a GA. In the first stage, features were filtered using the Fisher score, and in the second stage, GA was combined with the mRMR principle to select the final feature subset. However, the redundancy evaluation in this method focused only on pairwise interactions between features. Li et al.~\cite{li2022_24} introduced a binary individual search strategy-based bi-objective evolutionary algorithm (BIBE) . It removes irrelevant and redundant features using an improved Fisher score, and generates new individuals through binary crossover and binary mutation operators. Nevertheless, BIBE fails to effectively address the negative impact of duplicated solutions on feature selection performance. To mitigate the generation of duplicated feature subsets, Xu et al.~\cite{xu2020_25} proposed a duplication analysis based evolutionary algorithm (DAEA) for bi-objective feature selection. It evaluates the similarity between solutions in the search space using the Manhattan distance. Experimental results showed that the proposed duplicate elimination strategy significantly improved the performance of DAEA. Wang et al.~\cite{wang2021cyb_26} develop a multiobjective DE method using a clustering technique (MOCDE) for feature selection, which integrates K-means clustering to assist feature selection. Although MOCDE achieved promising results, it still tended to select some redundant features.

To further reduce redundancy in the selected feature subsets, Wang et al.~\cite{wang2022niching_27} developed a niche-based multiobjective differential evolution algorithm, which realized substantial improvements over baseline methods in terms of HV and IGD performance metrics. Xue et al.~\cite{xue2023_1} proposed a hybrid feature selection algorithm, which integrates the ReliefF-based multiobjective algorithm with both filter and wrapper methods. This hybrid approach enhances the effectiveness of solving feature selection problems. Ahadzadeh et al.~\cite{ahadzadeh2023_28} presented a simple, fast, and efficient feature selection algorithm (SFE), which achieved dimensionality reduction on high-dimensional datasets with low computational cost and memory usage. By combining it with PSO, they further proposed SFE-PSO, which significantly improved feature subset selection and enhanced classification accuracy. Song et al.~\cite{song2021_29} proposed a three-stage hybrid feature selection algorithm based on correlation-guided clustering and PSO. The method integrates feature filtering, fast clustering, and an improved integer PSO algorithm. Cheng et al.~\cite{cheng2021_30} proposed a steering-matrix-based multiobjective evolutionary algorithm, which significantly improved search efficiency and achieved high-quality feature subsets.
\section{Proposed Algorithm}\label{sec:Proposed}
\label{sec1}

This section first offers an overview of the proposed MODE-FS algorithm. It then describes the computation of the feature correlation weight matrix $Q$ and the redundancy index $A$, which are used to evaluate feature importance and redundancy, respectively. Finally, it elaborates on the core components of MODE-FS, including: WRBI, MSBIU, and FOAGM.

\subsection{Overall MODE-FS Framework}
\label{subsec1}

\begin{figure}
\centering
\includegraphics[width=1.0\textwidth]{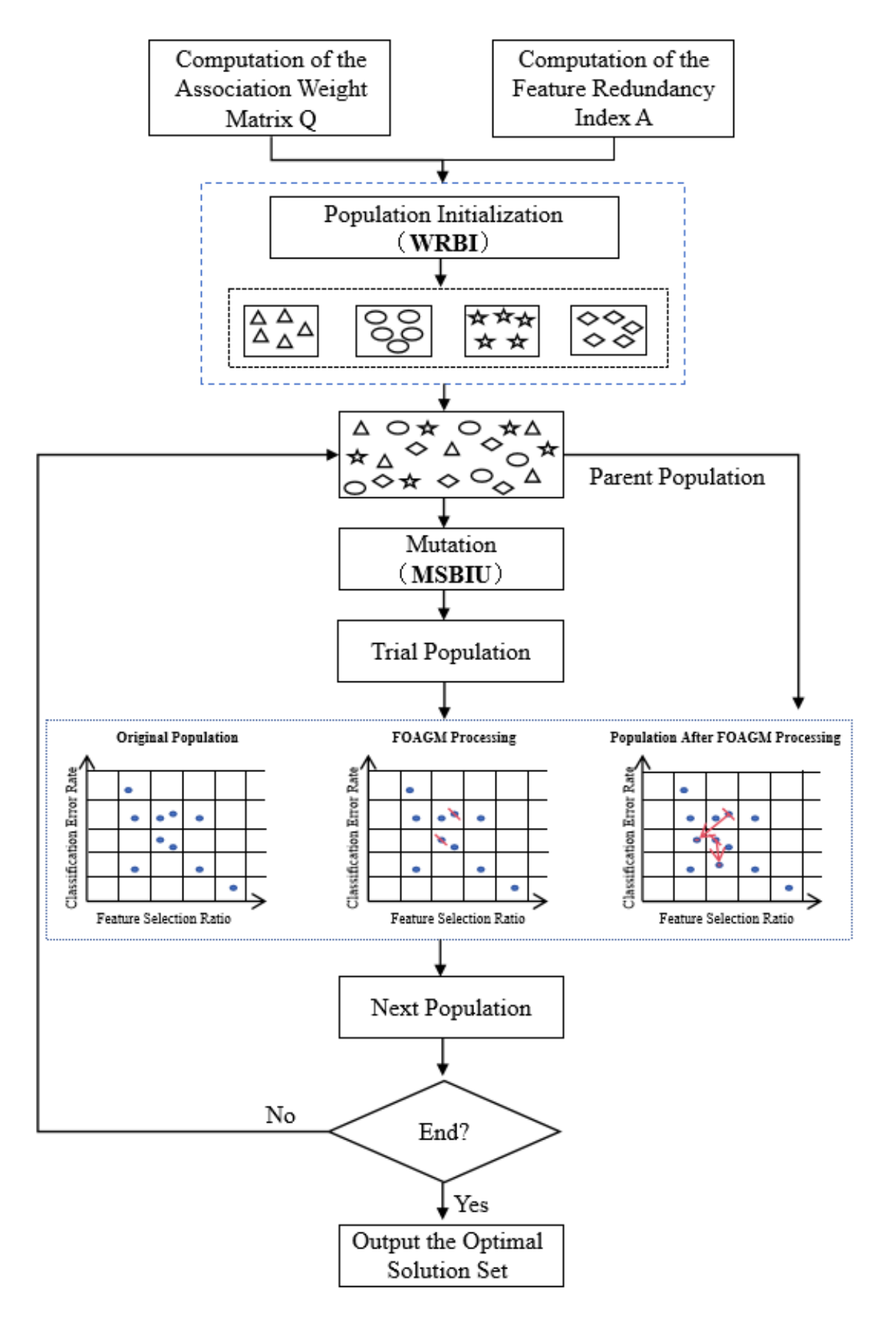}
\caption{The overall framework of the proposed MODE-FS.}
\label{fig1}
\end{figure}

This section presents the overall framework of the proposed MODE-FS algorithm. The optimization process is illustrated in Fig.~\ref{fig1}, and the complete procedure is outlined in Algorithm~\ref{alg1}. 
\begin{algorithm}[!t]
\caption{MODE-FS.}
\begin{algorithmic}[1]
\State \textbf{Input:} Population size $P$, maximum generations $MaxGen$, current generation $Ngen$
\State \textbf{Output:} The nondominated solutions in $P_M$
\State Initialize the population via WRBI (Algorithm~\ref{alg2})
\While{$Ngen \leq MaxGen$}
    \State Apply mutation and selection using MSBIU (Algorithm~\ref{alg3})
    \State Refine the population using FOAGM (Algorithm~\ref{alg4})
    \State $Ngen \gets Ngen + 1$
\EndWhile
\end{algorithmic}
\label{alg1}
\end{algorithm}

MODE-FS addresses multiobjective optimization in high-dimensional feature selection through multi-stage evolutionary computation, intending to enhance search efficiency, enhance solution distribution, and balance global exploration with local exploitation. The algorithm consists of three core modules: WRBI, MSBIU, and FOAGM. In the initialization phase, WRBI is employed to generate the initial population, utilizing the feature correlation weight matrix $Q$ to evaluate feature importance and the redundancy index matrix $A$ to assess feature redundancy. The population is divided into multiple subpopulations to enhance the diversity and distribution uniformity of the initial solutions, laying a robust groundwork for subsequent optimization. During the mutation and selection phase, MSBIU is applied to update the population. This mechanism integrates mean-shift adjustment with redundancy-aware filtering to dynamically refine feature subsets during mutation. Only features making a substantial contribution to the optimization objectives are retained, improving both solution quality and diversity, while enhancing global search ability. In the environmental selection phase, FOAGM is used to refine the distribution of solutions in the objective space. By adaptively partitioning the objective space into grids, FOAGM identifies densely populated regions and applies targeted feature optimization strategies to eliminate redundant features, thereby enhancing coverage, uniformity, and the quality of Pareto front solutions. The optimization process continues until the maximum number of generations $MaxGen$ is reached. All nondominated solutions are collected into the Pareto optimal set $P_M$. By integrating weight-guided initialization, redundancy-based filtering, and adaptive solution refinement, MODE-FS improves both global and local search capabilities, leading to more efficient feature selection, improved classification accuracy, and reduced redundancy. Experimental results demonstrate that MODE-FS consistently outperforms existing feature selection methods on multiple high-dimensional datasets. It effectively reduces the number of selected features while enhancing classification performance, offering a robust and efficient multiobjective optimization solution for high-dimensional feature selection tasks.

\subsection{Feature Correlation Weight Matrix $Q$ and Redundancy Index $A$}
\label{subsec1}
In the feature selection process, it is essential to evaluate the contribution of each feature to the classification target (label). Accordingly, this study employs an FCM to compute the weights of features relative to the label and to construct a feature correlation weight matrix $Q$ for guiding the feature selection process. FCM is a fuzzy logic-based graphical model, designed to represent the causal relationships between features and the target variable. It consists of feature nodes and a target node (label), where the feature nodes correspond to the input features $x_1$, $x_2$, $\dots$, $x_n$, and the target node represents the classification label $y$. By learning the influence weights $w_i$ between the features and the target label, FCM generates the feature correlation weight matrix $Q$:

\begin{equation}
Q = 
\begin{bmatrix}
w_1 \\
w_2 \\
\vdots \\
w_n
\end{bmatrix}
\label{eq14}
\end{equation}
where $w_i$ denotes the contribution of feature $x_i$ to the target variable $y$. The weights are computed through an iterative updating process based on FCM, as described in Eq.(\ref{eq15}).

\begin{equation}
a^{(t+1)}_y = f\left( \sum_{i=1}^{n} w_i a^{(t)}_i \right)
\label{eq15}
\end{equation}
where $a^{(t+1)}_y$ denotes the state value of the $j$-th node at the $(t+1)$-th iteration, and $f(\cdot)$ is an activation function (e.g., a threshold function or an S-shaped function) used to normalize the output to a specified range.
To assess the similarity between features, cosine similarity is employed as the similarity metric~\cite{kwak2021_31}. It is a widely adopted measure in high-dimensional space for quantifying the angular similarity between two non-zero vectors. The mathematical formulation is presented in Eq.(\ref{eq16}).

\begin{equation}
\text{Cosine Similarity}(\vec{x}_i, \vec{x}_j) = \frac{\vec{x}_i \cdot \vec{x}_j}{\|\vec{x}_i\| \|\vec{x}_j\|}
\label{eq16}
\end{equation}
where $\vec{x}_i$ and $\vec{x}_j$ represent two feature vectors, and $\|\vec{x}_i\|$ and $\|\vec{x}_j\|$ denote their respective Euclidean norms. In the feature selection process, cosine similarity is used to construct the feature redundancy index matrix $A$, which is defined as follows:
\begin{equation}
A = 
\begin{bmatrix}
A_{11} & A_{12} & \cdots & A_{1n} \\
A_{21} & A_{22} & \cdots & A_{2n} \\
\vdots & \vdots & \ddots & \vdots \\
A_{n1} & A_{n2} & \cdots & A_{nn}
\end{bmatrix}
\label{eq17}
\end{equation}
where each element $A_{ij}$ in matrix $A$ reflects the similarity between features $x_i$ and $x_j$. 
To effectively identify redundant features, a threshold $\tau$ is defined as the median value of all entries in matrix $A$, and is used as the decision criterion. 
When $A_{ij} > \tau$, it indicates that the corresponding feature pair exhibits high similarity and is considered potentially redundant. 
Conversely, when $A_{ij} \leq \tau$, the features are regarded as relatively independent, with low redundancy. 
The matrix $A$ is symmetric, i.e., $A_{ij} = A_{ji}$, and all diagonal elements satisfy $A_{ii} = 1$.

Under normal circumstances, the cosine similarity between different features is relatively low. However, during the feature selection process, if certain features have a substantial influence on the optimization objectives, cosine similarity can be leveraged to refine the feature selection strategy. In addition, a high cosine similarity typically indicates a greater degree of information redundancy between features. Therefore, prioritizing the removal of these redundant features enhances the diversity and effectiveness of the final selected feature subset.
\begin{algorithm}[!t]
\caption{WRBI.}
\begin{algorithmic}[1]
\State \textbf{Input:} Feature weight matrix $Q$, Feature redundancy index $A$, number of features $N$, population size $P$
\State \textbf{Output:} Initialized population ($Pop$)
\State Initialize $P_1$, $P_2$, $P_3$, $P_4$ as empty sets
\State $highWeight \gets$ Top 60\% of features based on $Q$
\State $lowWeight \gets$ Bottom 40\% of features based on $Q$
\For{$i = 1$ to $P_1$ size}
    \State Select a random subset of features from $highWeight$
    \State Add the selected features to the $i$-th individual of $P_1$
\EndFor
\For{$i = 1$ to $P_2$ size}
    \State Select a random subset of features from $lowWeight$
    \State Add the selected features to the $i$-th individual of $P_2$
\EndFor
\For{$i = 1$ to $P_3$ size}
    \State Select high-weight features from $highWeight$
    \State Remove redundant features based on $A$
    \State Add the selected features to the $i$-th individual of $P_3$
\EndFor
\For{$i = 1$ to $P_4$ size}
    \State Select high-weight features from $highWeight$
    \State Add the selected features to the $i$-th individual of $P_4$
\EndFor
\State $Pop \gets P_1 \cup P_2 \cup P_3 \cup P_4$
\While{$|Pop| < P$}
    \State Select a random subset of features from $highWeight$
    \State Add the selected features to $Pop$
\EndWhile
\end{algorithmic}
\label{alg2}
\end{algorithm}
\begin{figure}[t]
    \centering
    \begin{subfigure}{0.45\textwidth}
        \centering
        \includegraphics[width=\linewidth]{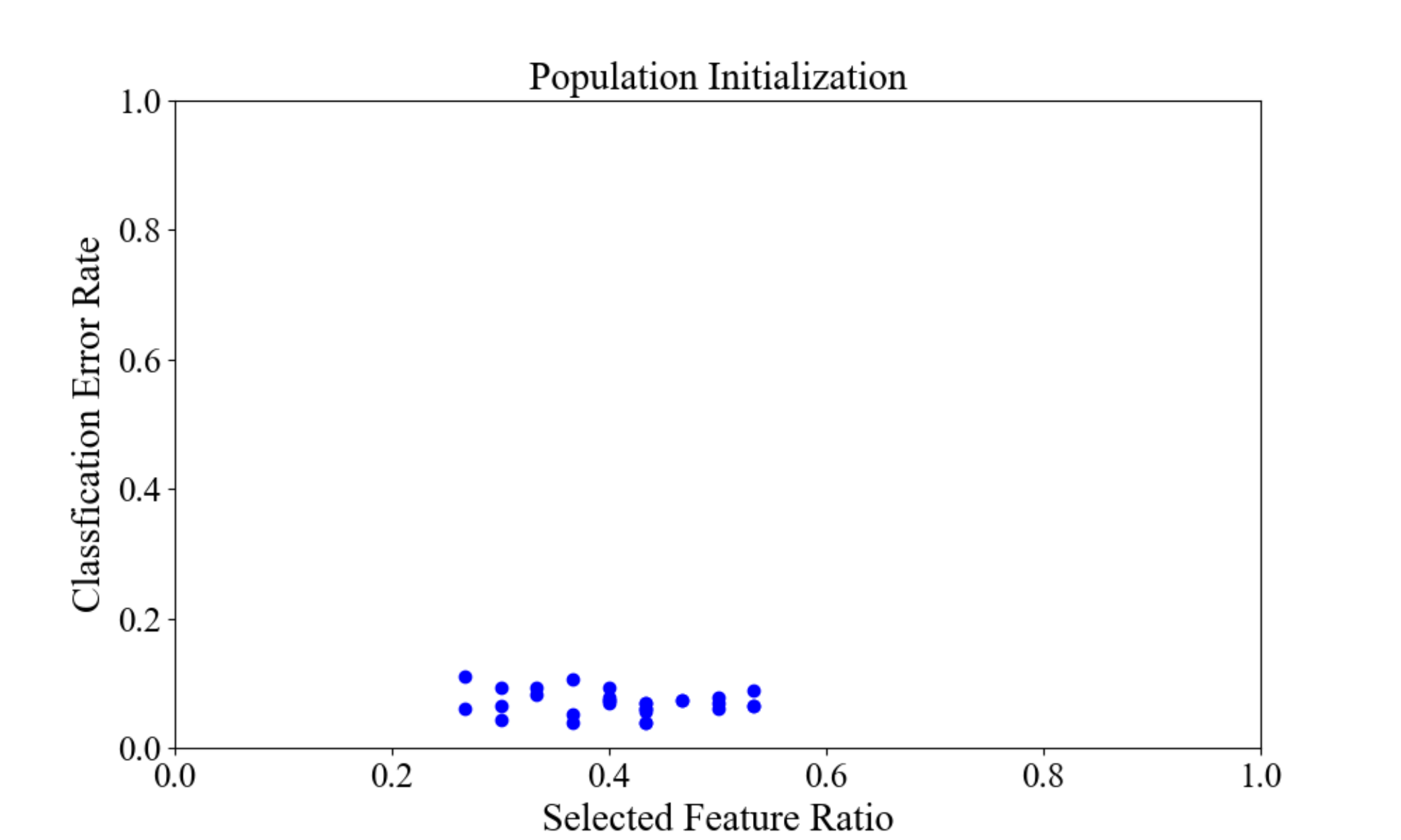} 
    \end{subfigure}
    \hfill
    \begin{subfigure}{0.45\textwidth}
        \centering
        \includegraphics[width=\linewidth]{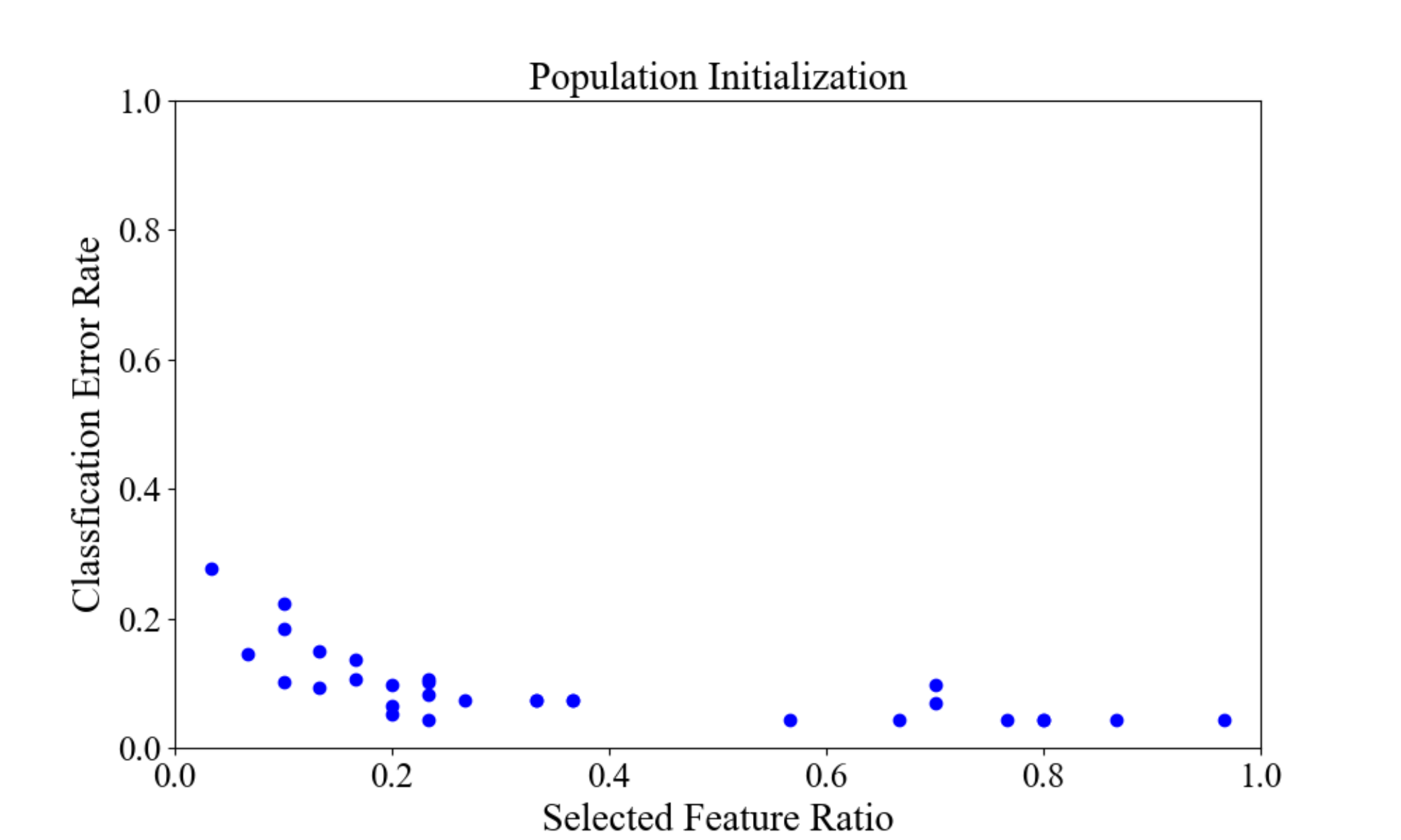} 
    \end{subfigure}
    \caption{Two populations generated by the proposed initialization strategy and the random initialization strategy. The population on the left is generated by the random initialization strategy, while the population on the right is generated by the proposed initialization strategy.}
    \label{fig2}
\end{figure}
\subsection{Weighted-Redundancy Balanced Initialization(WRBI)}
In this section, we propose a WRBI strategy. This approach first computes the feature correlation weight matrix $Q$ using the FCM method to assess the importance of each feature. Additionally, cosine similarity is employed to calculate the redundancy index matrix $A$, which quantifies the redundancy between features. Based on this information, the population is evenly divided into four subpopulations: $P_1$, $P_2$, $P_3$, and $P_4$. For each individual in these subpopulations, features that meet the corresponding criteria are randomly selected to form a candidate solution. Specifically, $P_1$ consists of individuals generated from high-weight features, formed after eliminating the bottom $40$\% of low-weight features. $P_2$ contains low-quality solutions, constructed from the remaining features without considering redundancy, following the removal of the bottom $40$\% of low-weight features. $P_3$ represents an elite subpopulation composed of high-weight and redundancy-reduced features, while $P_4$ contains elite individuals formed from high-weight features without redundancy filtering. These four subpopulations are then merged to form the initial population. The detailed procedure of WRBI is presented in Algorithm~\ref{alg2}. Furthermore, the distributions of populations generated by WRBI and by random initialization are illustrated in Fig.~\ref{fig2}. Compared to the randomly initialized population, the WRBI-generated population demonstrates superior convergence and diversity, which validates the effectiveness of the WRBI strategy.

\subsection{Mutation-Selection Based Individual Update(MSBIU)}

In this section, we introduce a MSBIU method to optimize the feature selection process of population individuals. Specifically, for each individual $X_i$ in the population, a mutation vector $v_i$ is generated using Eq.(\ref{eq7}), with the scaling factor $F$ set to 0.5. Then, a mean shift operation is performed on the feature correlation weight matrix $Q$, resulting in the shifted weight matrix $Q_1$, as defined in Eq.(\ref{eq18}).

\begin{equation}
Q_1[j] = Q[i] - \text{mean}(Q), \quad \forall j \in \{1,2,\dots,N\}
\label{eq18}
\end{equation}

In this process, the values of $v'$ may include both positive and negative values. We retain only the features for which $v'>0$, ensuring that the selected features are consistent with the directionality imposed by $Q_1$. This strategy not only ensures that the mutation vector reflects the guiding influence of high-weight features but also prevents the introduction of ineffective features due to negative values, thereby enhancing the overall rationality and effectiveness of the feature selection process. During feature selection, the decision to include feature $j$ in the new individual $X_{\text{new}}$ is made based on the feature's redundancy value $A[j]$, the information from the mutation vector $v'$, and the position of feature $j$ in the current individual $X_i$. Specifically, if the redundancy value $A[j]$ of feature $j$ is below the threshold $\tau$ and $v'[j]>0$, feature $j$ is added to the new individual $X_{\text{new}}$; otherwise, it is excluded.

After updating the features, the new individual $X_{\text{new}}$ is reinitialized: if feature $j$ is selected, its value is randomly assigned between 0.6 and 1; otherwise, it is assigned a value between 0 and 0.6. The objective value of $X_{\text{new}}$ is then calculated and compared with that of the current individual $X_i$. If $X_{\text{new}}$ is superior in terms of nondominated sorting, it replaces $X_i$ in the next generation; otherwise, $X_i$ is retained.

\begin{figure}[t]
\centering
\includegraphics[width=0.75\textwidth]{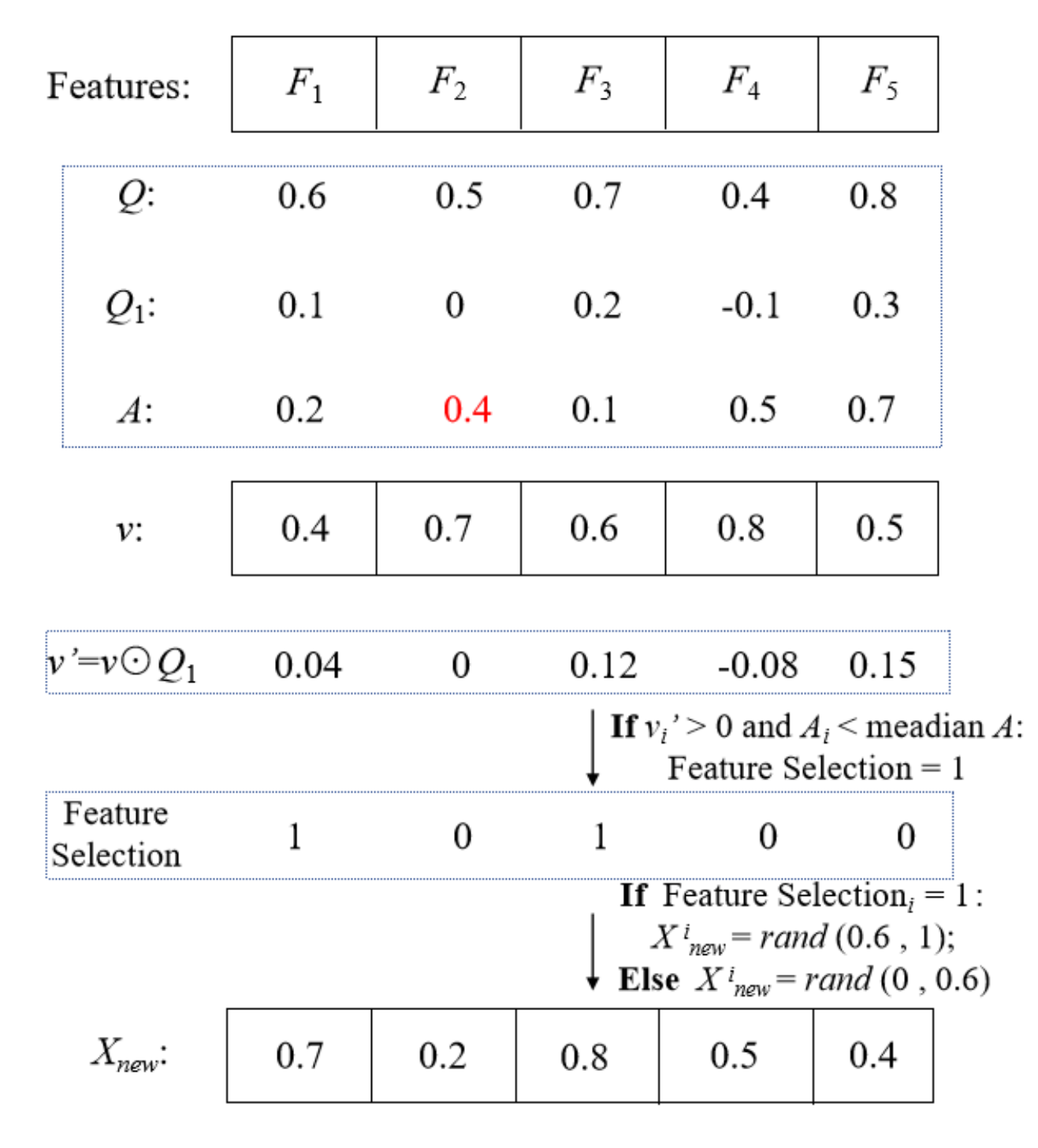}  
\caption{Detailed Processing Steps of MSBIU.}
\label{fig3}
\end{figure}

Fig.~\ref{fig3} gives a clear example of this process. Initially, five features, $F_1$, $F_2$, $F_3$, $F_4$, and $F_5$ are considered, and their corresponding matrices $Q$, $Q_1$, and $A$ are calculated, where the red areas in $A$ indicate the threshold $\tau$, and the symbol $\odot$ denotes the Hadamard product (element-wise multiplication). Each mutation vector $v_i$ is then processed, ultimately leading to the generation of the new individual $X_{\text{new}}$. Through this process, individual diversity is maintained and optimized, thereby enhancing the overall search performance of the population. The detailed procedure of the MSBIU algorithm is described in Algorithm~\ref{alg3}.

\begin{algorithm}[t]
\caption{MSBIU.}
\begin{algorithmic}[1]
\State \textbf{Input:} Population, weight matrix $Q$, redundancy index $A$, threshold $\tau$, scaling factor $F=0.5$
\State \textbf{Output:} New population (\textit{NewPopulation})
\State Initialize \textit{NewPopulation} as an empty set
\For{$X_i \in$ Population}
    \State Randomly select three individuals $r_1, r_2, r_3 \in$ Population, where $r_1 \neq r_2 \neq r_3 \neq X_i$
    \State Generate the mutation vector according to Eq.(\ref{eq7})
    \State Apply mean shift to the weight matrix $Q$ to obtain $Q_1$
    \State Adjust the mutation vector using $Q_1$: $v' = v \odot Q_1$
    \State Initialize $X_{\text{new}}$ as an empty set
    \For{each feature position $j$ in $X_i$}
        \If{$A[j] < \tau$ and $v'[j] > 0$}
            \State Add feature $j$ to $X_{\text{new}}$
        \Else
            \State Do not add feature $j$
        \EndIf
    \EndFor
    \State Calculate the objective value of $X_{\text{new}}$
    \If{$X_{\text{new}}$ is better than $X_i$ (based on nondominated sorting)}
        \State Add $X_{\text{new}}$ to \textit{NewPopulation}
    \Else
        \State Add $X_i$ to \textit{NewPopulation}
    \EndIf
\EndFor
\end{algorithmic}
\label{alg3}
\end{algorithm}

\subsection{Feature-Optimized Adaptive Grid Mechanism(FOAGM)}
\begin{algorithm}[!t]
\caption{FOAGM.}
\begin{algorithmic}[1]
\State \textbf{Input:} Population, objective values, feature weight matrix $Q$, redundancy index $A$, population size $P$
\State \textbf{Output:} Optimized population
\State Divide the objective space into a $P\times P$ grid $G$
\For{each grid $G$}
    \State Count the number of individuals: $\text{count}(G)$
    \If{$\text{count}(G) > \rho_{\text{threshold}}$}
        \State Mark $G$ as a dense grid
    \EndIf
\EndFor
\For{each dense grid}
    \State Subdivide the grid into smaller subgrids
    \State Redistribute individuals into new subgrids
    \State Recount the number of individuals in each subgrid
    \If{a subgrid remains dense}
        \State Repeat steps 11-13
    \Else
        \State Stop subdivision
    \EndIf
\EndFor
\For{each individual in dense grids}
    \State Perform nondominated sorting with classification error rate as priority
    \For{each inferior individual}
        \State Identify the feature $j_{\min} = \arg\min(Q[j])$
        \State Remove the feature $j_{\min}$
        \State Select and add a feature $j_{\text{new}} = \arg\max\left(\frac{Q[j]}{A[j]}\right)$ where $j \notin$ individual
    \EndFor
\EndFor
\end{algorithmic}
\label{alg4}
\end{algorithm}
To further enhance the optimization efficiency of the population and the effectiveness of feature selection, an adaptive grid mechanism is introduced into the optimization framework, proposing the FOAGM. The detailed procedure of the FOAGM algorithm is described in Algorithm~\ref{alg4}. FOAGM enables dynamic monitoring and adjustment of the population distribution, thereby balancing solution diversity and performance.

During each generation of the evolutionary process, the population is first mapped into the objective space. Then, a dynamic grid is applied to partition the population into $P \times P$ grid cells $G\{x,y\}$, as defined in Eq.(\ref{eq19}).

\begin{equation}
G(x,y) = \{f_i \,|\, f_{i1} \in [x, x+\Delta_x), f_{i2} \in [y, y+\Delta_y)\}
\label{eq19}
\end{equation}
where
\[
\Delta_x = \frac{\max(f_1) - \min(f_1)}{n}, \quad
\Delta_y = \frac{\max(f_2) - \min(f_2)}{n}.
\]

The number of individuals within each grid cell is counted. If the number of individuals in a grid cell exceeds a predefined threshold, the cell is identified as a "dense grid". For individuals located in dense grids, a nondominated sorting procedure is performed to select the nondominated solutions for the next generation. If all individuals within the grid are nondominated, the 
classification error rate is used as the secondary criterion, prioritizing the retention of individuals with lower error rates in accordance with the objectives of the feature selection task.

\begin{figure}[t]
\centering
\includegraphics[width=1.0\textwidth]{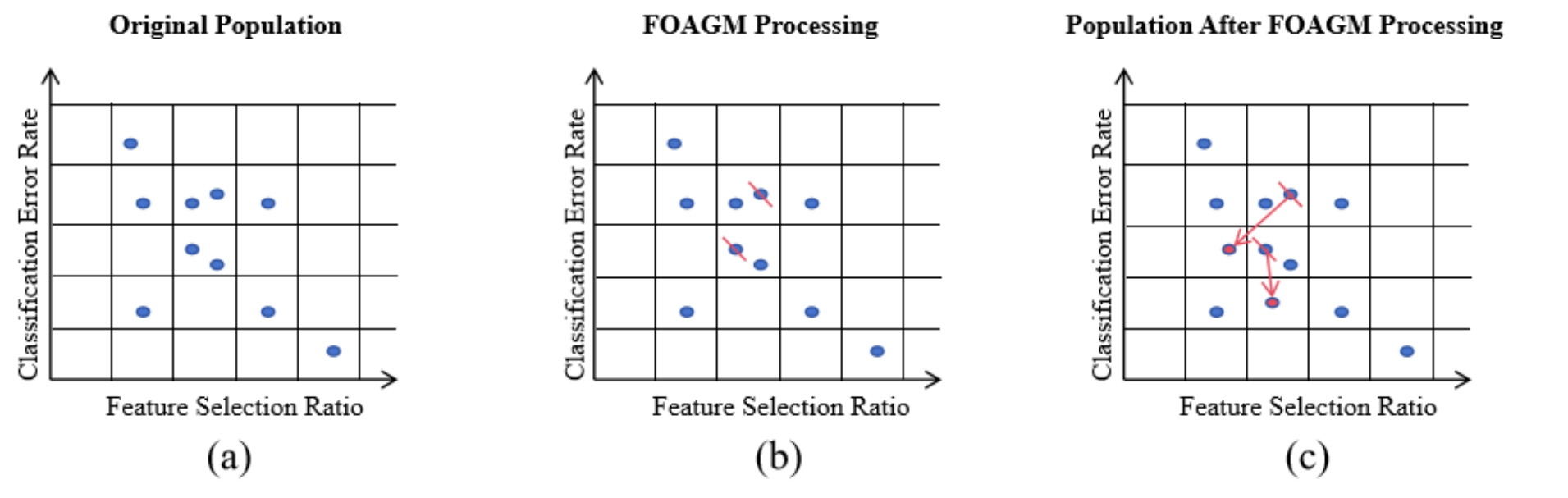}
\caption{FOAGM Processing Procedure.}
\label{fig4}
\end{figure}

To dynamically adapt to changes in population distribution, dense grids are further subdivided into smaller subgrids, and individuals are reassigned accordingly. Further subdivision continues until a sufficiently sparse distribution is achieved, thereby ensuring a rational allocation of computational resources. For global feature selection optimization, the algorithm leverages both the feature correlation weight matrix $Q$ and the redundancy index matrix $A$ to refine inferior individuals in dense grids. Specifically, features with low weights and high redundancy are eliminated, while features with high weights and low redundancy are incorporated, thereby maintaining the structural consistency of individuals while enhancing their fitness. As illustrated in Fig.~\ref{fig4}, two grid regions with significantly higher densities than others can be clearly observed in Fig.~\ref{fig4}(a). For these dense grids, the FOAGM method is applied: inferior individuals are filtered out and refined, and the remaining individuals are redistributed into relatively sparse regions. This process not only reduces individual density within dense grids and mitigates local crowding but also significantly improves the comprehensive standard of individuals and enhances population diversity, thereby facilitating subsequent optimization processes.

\section{Experiments}\label{sec:Experiments}
\subsection{Datasets}

In this study, $11$ representative datasets were selected from the UCI Machine Learning Repository~\cite{dua2017_32}, spanning domains such as healthcare, bioinformatics, and signal processing, to comprehensively evaluate the performance of the proposed method on multiobjective feature selection tasks. These datasets exhibit considerable diversity in feature dimensionality, sample size, and number of classes, thus providing robust and varied evaluation benchmarks. The characteristics of the datasets are summarized in Table~\ref{tab1}. 
\begin{table}[H]
\centering
\begin{tabular}{c l c c c}
\hline
No. & Dataset &  Features &  Instances &  Classes \\
\hline
1 & Segment    & 19   & 2310 & 7 \\
2 & WBCD       & 30   & 569  & 2 \\
3 & Ionosphere & 34   & 351  & 2 \\
4 & Sonar      & 60   & 208  & 2 \\
5 & Hillvalley & 100  & 1212 & 2 \\
6 & Musk1      & 166  & 476  & 2 \\
7 & Semeion    & 256  & 1593 & 10 \\
8 & Arrhythmia & 279  & 452  & 16 \\
9 & Toxicity   & 1203 & 171  & 2 \\
10 & SRBCT     & 2308 & 83   & 4 \\
11 & DLBCL     & 5469 & 77   & 2 \\
\hline
\end{tabular}
\caption{Information of Datasets.}
\label{tab1}
\end{table}
To further evaluate the robustness of the method under high-dimensional and small-sample conditions, three additional datasets (Toxicity, SRBCT, and DLBCL) were included. These datasets are characterized by extremely high dimensionality and limited sample sizes, presenting a substantial obstacle for feature selection algorithms. In the experimental setup, each classification dataset is randomly divided into a training subset and a testing subset with proportions of approximately $60$\% and $40$\%, respectively. During the training phase, a K-Nearest Neighbor (KNN) classifier was employed to evaluate the classification performance based on the selected feature subsets, while the testing set was used for final performance verification. This partitioning strategy ensures a reliable and objective evaluation of the feature subsets' effectiveness in classification tasks.

\subsection{Benchmark Techniques and Parameter Settings}

To assess the performance of the proposed algorithm, a comparative analysis is conducted between MODE-FS and five FS methods. They are NSGA-II~\cite{deb2002_7}, MOEA/D~\cite{zhang2007_8}, MFFS~\cite{jiao2023_4}, MOCDE~\cite{yu2025_17}, and CMODE~\cite{pan2022_6}. Among them, NSGA-II and MOEA/D are two representative EMO algorithms, whereas MFFS, MOCDE, and CMODE are three popular EMO-based feature selection approaches.

Each feature selection algorithm was independently executed 30 times on each training dataset. For all datasets, the maximum number of maximum iterations ($T$) was set to $100 \times P$, where $P$ denotes the population size. The population size was set proportional to the number of original features in each dataset, with an upper limit of 200. This configuration ensures a comprehensive and standardized evaluation of algorithm performance in high-dimensional feature spaces. 

For real-coded representation, the feature selection threshold $\theta$ was set to $0.6$. Typically, $\theta$ is chosen within the range $[0.5, 0.7]$ for feature selection tasks~\cite{tran2019_33}, with variations within this range generally having minimal impact on the selection outcomes. In this study, $\theta=0.6$ was adopted to moderately favor the selection of fewer features during the early stages of the evolutionary process. Specifically, in the decoding process, a feature is selected if its corresponding real value is greater than or equal to $\theta$, i.e., $X_{i,j} \geq \theta$. All algorithms were terminated once the predefined maximum number of fitness evaluations was reached, ensuring consistency across all comparative methods.

To assess the overall performance of the six EMO algorithms, the hypervolume (HV) indicator is adopted~\cite{zitzler1999_34}. It measures the volume of the objective space that is weakly dominated by the obtained nondominated solutions. The mathematical expression of HV is provided in Eq.(\ref{eq20}).
\begin{equation}
\text{HV}(S, r_p) = \mathcal{L} \left( \bigcup_{i \in S} \left\{ y \mid x \prec y \prec r_p \right\} \right)
\label{eq20}
\end{equation}
where $S$ denotes the set of nondominated solutions, $r_p$ is the reference point in the objective space, and $\mathcal{L}(\cdot)$ represents the Lebesgue measure used to compute the hypervolume. The notation $x \prec y$ indicates that solution $x$ dominates $y$. The reference point $r_p$ for HV computation was set to $(1, 1)$, following the recommendation in~\cite{xu2020gecco_35}. The detailed parameter settings are summarized in Table~\ref{tab2}.
\begin{table}[H]
\centering
\begin{tabular}{c c}
\hline
Parameters & Settings \\
\hline
$P$ (population size) & \# Features ($N$) and restriction to 200 \\
$T$ (maximum iterations) & $100 \times P$ \\
$K$ (KNN) & 5 \\
$\theta$ & 0.6 \\
$r_p$ & [1,1] \\
\hline
\end{tabular}
\caption{Parameter Setting.}
\label{tab2}
\end{table}

In addition, the Inverted Generational Distance (IGD) is employed to assess both the convergence and distribution of the solution set by measuring its proximity to the true Pareto front~\cite{bosman2003_36}. Specifically, IGD is defined as the average Euclidean distance from each point on the reference Pareto front to its nearest solution in the obtained nondominated set, as defined in Eq.(\ref{eq21}).
\begin{equation}
\mathrm{IGD}(S, P^*) = \frac{1}{|P^*|} \sum_{v \in P^*} \min_{x \in S} \|x - v\|
\label{eq21}
\end{equation}
where $S$ denotes the set of nondominated solutions obtained by the algorithm, and $P^*$ represents a set of uniformly distributed reference points sampled from the true Pareto front. A lower IGD value indicates that the obtained solution set is both closer to and more evenly distributed along the true front.

Since the true Pareto front is typically unknown in feature selection problems, to compute IGD, all nondominated solutions obtained by the comparative algorithms on the test sets were first aggregated into a joint population. Subsequently, the nondominated solutions from this joint population were treated as an approximation of the “true” Pareto front.
\section{Results}\label{sec:Results}
\subsection{Analysis of Feature Selection Performance}

Tables~\ref{tab3} and~\ref{tab4} present the experimental results comparing MODE-FS against five highly esteemed MOEA-based feature selection methods in terms of HV and IGD, respectively. The results clearly demonstrate that MODE-FS exhibits considerable overall superiority in multiobjective feature selection tasks. Specifically, MODE-FS consistently attained substantial improvements across most datasets in both HV and IGD, showing notable advantages in the uniformity of solution distribution and the quality of convergence. Although MFFS and NSGA-II achieved locally optimal IGD values on a few individual datasets, MODE-FS demonstrated more consistent and broadly applicable performance across all datasets.

\begin{table*}[t]
\centering
\resizebox{\textwidth}{!}{

\begin{tabular}{lcccccc}
\toprule
Dataset & NSGA-II & MOEA/D & MFFS & MOCDE & CMODE & \textbf{MODE-FS} \\
\midrule
Segment & \begin{tabular}{@{}c@{}}6.629e-01\\$\pm$1.564e-02\end{tabular} & \begin{tabular}{@{}c@{}}7.324e-01\\$\pm$1.242e-03\end{tabular} & \begin{tabular}{@{}c@{}}7.788e-01\\$\pm$3.487e-03\end{tabular} & \begin{tabular}{@{}c@{}}7.695e-01\\$\pm$8.421e-03\end{tabular} & \begin{tabular}{@{}c@{}}6.302e-01\\$\pm$1.067e-02\end{tabular} & \textbf{\begin{tabular}{@{}c@{}}7.841e-01\\$\pm$1.787e-02\end{tabular}} \\
WBCD & \begin{tabular}{@{}c@{}}8.300e-01\\$\pm$4.831e-03\end{tabular} & \begin{tabular}{@{}c@{}}7.300e-01\\$\pm$5.067e-03\end{tabular} & \begin{tabular}{@{}c@{}}8.900e-01\\$\pm$1.555e-02\end{tabular} & \begin{tabular}{@{}c@{}}8.500e-01\\$\pm$3.788e-02\end{tabular} & \begin{tabular}{@{}c@{}}6.600e-01\\$\pm$2.931e-02\end{tabular} & \textbf{\begin{tabular}{@{}c@{}}9.152e-01\\$\pm$1.793e-03\end{tabular}} \\
Ionosphere & \begin{tabular}{@{}c@{}}7.600e-01\\$\pm$6.622e-03\end{tabular} & \begin{tabular}{@{}c@{}}6.000e-01\\$\pm$2.599e-02\end{tabular} & \begin{tabular}{@{}c@{}}8.600e-01\\$\pm$1.625e-02\end{tabular} & \begin{tabular}{@{}c@{}}7.800e-01\\$\pm$2.264e-02\end{tabular} & \begin{tabular}{@{}c@{}}4.700e-01\\$\pm$9.462e-03\end{tabular} & \textbf{\begin{tabular}{@{}c@{}}8.765e-01\\$\pm$2.491e-02\end{tabular}} \\
Sonar & \begin{tabular}{@{}c@{}}6.300e-01\\$\pm$2.711e-02\end{tabular} & \begin{tabular}{@{}c@{}}5.600e-01\\$\pm$8.112e-02\end{tabular} & \begin{tabular}{@{}c@{}}6.900e-01\\$\pm$3.412e-02\end{tabular} & \begin{tabular}{@{}c@{}}7.400e-01\\$\pm$8.115e-03\end{tabular} & \begin{tabular}{@{}c@{}}3.900e-01\\$\pm$1.870e-02\end{tabular} & \textbf{\begin{tabular}{@{}c@{}}8.370e-01\\$\pm$3.935e-02\end{tabular}} \\
Hillvalley & \begin{tabular}{@{}c@{}}4.300e-01\\$\pm$2.130e-02\end{tabular} & \begin{tabular}{@{}c@{}}3.400e-01\\$\pm$8.314e-02\end{tabular} & \begin{tabular}{@{}c@{}}5.300e-01\\$\pm$1.876e-02\end{tabular} & \begin{tabular}{@{}c@{}}4.900e-01\\$\pm$1.217e-02\end{tabular} & \begin{tabular}{@{}c@{}}2.800e-01\\$\pm$2.421e-03\end{tabular} & \textbf{\begin{tabular}{@{}c@{}}5.541e-01\\$\pm$1.348e-03\end{tabular}} \\
Musk1 & \begin{tabular}{@{}c@{}}6.826e-01\\$\pm$4.684e-03\end{tabular} & \begin{tabular}{@{}c@{}}6.242e-01\\$\pm$1.642e-02\end{tabular} & \begin{tabular}{@{}c@{}}7.417e-01\\$\pm$3.375e-02\end{tabular} & \begin{tabular}{@{}c@{}}8.186e-01\\$\pm$2.718e-02\end{tabular} & \begin{tabular}{@{}c@{}}4.601e-01\\$\pm$1.550e-02\end{tabular} & \textbf{\begin{tabular}{@{}c@{}}8.667e-01\\$\pm$2.127e-02\end{tabular}} \\
Semeion & \begin{tabular}{@{}c@{}}5.888e-01\\$\pm$1.842e-02\end{tabular} & \begin{tabular}{@{}c@{}}6.172e-01\\$\pm$1.377e-03\end{tabular} & \begin{tabular}{@{}c@{}}6.512e-01\\$\pm$2.375e-02\end{tabular} & \begin{tabular}{@{}c@{}}5.888e-01\\$\pm$4.931e-03\end{tabular} & \begin{tabular}{@{}c@{}}5.224e-01\\$\pm$1.114e-02\end{tabular} & \textbf{\begin{tabular}{@{}c@{}}8.294e-01\\$\pm$3.855e-02\end{tabular}} \\
Arrhythmia & \begin{tabular}{@{}c@{}}4.838e-01\\$\pm$6.666e-03\end{tabular} & \begin{tabular}{@{}c@{}}4.209e-01\\$\pm$1.591e-02\end{tabular} & \textbf{\begin{tabular}{@{}c@{}}6.636e-01\\$\pm$1.168e-02\end{tabular}} & \begin{tabular}{@{}c@{}}5.630e-01\\$\pm$6.384e-05\end{tabular} & \begin{tabular}{@{}c@{}}3.162e-01\\$\pm$1.545e-03\end{tabular} & \begin{tabular}{@{}c@{}}5.890e-01\\$\pm$2.501e-02\end{tabular} \\
Toxicity & \begin{tabular}{@{}c@{}}4.634e-01\\$\pm$1.619e-02\end{tabular} & \begin{tabular}{@{}c@{}}4.084e-01\\$\pm$1.828e-02\end{tabular} & \textbf{\begin{tabular}{@{}c@{}}7.373e-01\\$\pm$3.155e-02\end{tabular}} & \begin{tabular}{@{}c@{}}6.448e-01\\$\pm$1.372e-02\end{tabular} & \begin{tabular}{@{}c@{}}3.276e-01\\$\pm$8.452e-03\end{tabular} & \begin{tabular}{@{}c@{}}6.621e-01\\$\pm$2.322e-02\end{tabular} \\
SRBCT & \begin{tabular}{@{}c@{}}5.763e-01\\$\pm$1.485e-02\end{tabular} & \begin{tabular}{@{}c@{}}5.822e-01\\$\pm$7.551e-04\end{tabular} & \begin{tabular}{@{}c@{}}9.307e-01\\$\pm$2.680e-02\end{tabular} & \begin{tabular}{@{}c@{}}9.167e-01\\$\pm$2.278e-02\end{tabular} & \begin{tabular}{@{}c@{}}5.084e-01\\$\pm$1.015e-03\end{tabular} & \textbf{\begin{tabular}{@{}c@{}}9.332e-01\\$\pm$2.194e-02\end{tabular}} \\
DLBCL & \begin{tabular}{@{}c@{}}5.719e-01\\$\pm$2.010e-02\end{tabular} & \begin{tabular}{@{}c@{}}6.081e-01\\$\pm$8.520e-04\end{tabular} & \begin{tabular}{@{}c@{}}9.824e-01\\$\pm$1.206e-02\end{tabular} & \begin{tabular}{@{}c@{}}9.450e-01\\$\pm$6.390e-03\end{tabular} & \begin{tabular}{@{}c@{}}5.512e-01\\$\pm$2.146e-02\end{tabular} & \textbf{\begin{tabular}{@{}c@{}}9.974e-01\\$\pm$1.352e-02\end{tabular}} \\
\bottomrule
\end{tabular}
} 
\caption{Comparison of HV values obtained by NSGA-II, MOEA/D, MFFS, MOCDE, CMODE, and MODE-FS on the test datasets.}
\label{tab3}
\end{table*}

\begin{table*}[t]
\centering
\resizebox{\textwidth}{!}{
\begin{tabular}{lcccccc}
\toprule
Dataset & NSGA-II & MOEA/D & MFFS & MOCDE & CMODE & \textbf{MODE-FS} \\
\midrule
Segment & \textbf{\begin{tabular}{@{}c@{}}6.339e-02\\$\pm$5.696e-03\end{tabular}} & \begin{tabular}{@{}c@{}}1.552e-01\\$\pm$1.142e-03\end{tabular} & \begin{tabular}{@{}c@{}}1.081e-01\\$\pm$2.563e-03\end{tabular} & \begin{tabular}{@{}c@{}}7.839e-02\\$\pm$1.966e-03\end{tabular} & \begin{tabular}{@{}c@{}}1.739e-01\\$\pm$1.358e-03\end{tabular} & \begin{tabular}{@{}c@{}}1.034e-01\\$\pm$2.402e-03\end{tabular} \\
WBCD & \begin{tabular}{@{}c@{}}5.822e-02\\$\pm$3.148e-03\end{tabular} & \begin{tabular}{@{}c@{}}5.899e-02\\$\pm$2.315e-03\end{tabular} & \begin{tabular}{@{}c@{}}8.280e-02\\$\pm$1.146e-03\end{tabular} & \begin{tabular}{@{}c@{}}1.128e-01\\$\pm$2.979e-03\end{tabular} & \begin{tabular}{@{}c@{}}1.724e-01\\$\pm$9.816e-03\end{tabular} & \textbf{\begin{tabular}{@{}c@{}}5.336e-02\\$\pm$2.269e-03\end{tabular}} \\
Ionosphere & \begin{tabular}{@{}c@{}}1.371e-02\\$\pm$2.176e-03\end{tabular} & \begin{tabular}{@{}c@{}}1.581e-01\\$\pm$1.822e-02\end{tabular} & \textbf{\begin{tabular}{@{}c@{}}5.908e-03\\$\pm$2.769e-03\end{tabular}} & \begin{tabular}{@{}c@{}}8.546e-02\\$\pm$2.084e-03\end{tabular} & \begin{tabular}{@{}c@{}}3.347e-01\\$\pm$2.063e-03\end{tabular} & \begin{tabular}{@{}c@{}}3.132e-02\\$\pm$2.249e-03\end{tabular} \\
Sonar & \begin{tabular}{@{}c@{}}7.211e-02\\$\pm$1.608e-03\end{tabular} & \begin{tabular}{@{}c@{}}2.363e-01\\$\pm$4.593e-03\end{tabular} & \begin{tabular}{@{}c@{}}8.345e-02\\$\pm$4.818e-03\end{tabular} & \begin{tabular}{@{}c@{}}1.059e-01\\$\pm$1.909e-02\end{tabular} & \begin{tabular}{@{}c@{}}3.699e-01\\$\pm$1.166e-02\end{tabular} & \textbf{\begin{tabular}{@{}c@{}}8.161e-02\\$\pm$3.446e-03\end{tabular}} \\
Hillvalley & \begin{tabular}{@{}c@{}}1.223e-01\\$\pm$1.062e-02\end{tabular} & \begin{tabular}{@{}c@{}}2.759e-01\\$\pm$4.345e-03\end{tabular} & \begin{tabular}{@{}c@{}}4.916e-02\\$\pm$2.066e-03\end{tabular} & \begin{tabular}{@{}c@{}}1.081e-01\\$\pm$4.162e-03\end{tabular} & \begin{tabular}{@{}c@{}}3.930e-01\\$\pm$4.117e-03\end{tabular} & \textbf{\begin{tabular}{@{}c@{}}4.888e-02\\$\pm$1.696e-02\end{tabular}} \\
Musk1 & \begin{tabular}{@{}c@{}}1.559e-01\\$\pm$1.259e-02\end{tabular} & \begin{tabular}{@{}c@{}}2.737e-01\\$\pm$1.416e-03\end{tabular} & \begin{tabular}{@{}c@{}}1.968e-02\\$\pm$6.702e-03\end{tabular} & \begin{tabular}{@{}c@{}}9.058e-02\\$\pm$6.484e-03\end{tabular} & \begin{tabular}{@{}c@{}}4.101e-01\\$\pm$2.114e-03\end{tabular} & \textbf{\begin{tabular}{@{}c@{}}1.868e-02\\$\pm$4.251e-03\end{tabular}} \\
Semeion & \begin{tabular}{@{}c@{}}1.930e-01\\$\pm$3.909e-03\end{tabular} & \begin{tabular}{@{}c@{}}2.561e-01\\$\pm$1.370e-02\end{tabular} & \textbf{\begin{tabular}{@{}c@{}}1.265e-02\\$\pm$4.885e-03\end{tabular}} & \begin{tabular}{@{}c@{}}6.834e-02\\$\pm$1.013e-03\end{tabular} & \begin{tabular}{@{}c@{}}3.235e-01\\$\pm$4.926e-03\end{tabular} & \begin{tabular}{@{}c@{}}4.217e-02\\$\pm$6.058e-03\end{tabular} \\
Arrhythmia & \begin{tabular}{@{}c@{}}2.071e-01\\$\pm$3.105e-03\end{tabular} & \begin{tabular}{@{}c@{}}3.297e-01\\$\pm$2.946e-03\end{tabular} & \begin{tabular}{@{}c@{}}6.577e-02\\$\pm$3.222e-03\end{tabular} & \begin{tabular}{@{}c@{}}1.524e-01\\$\pm$1.829e-02\end{tabular} & \begin{tabular}{@{}c@{}}4.349e-01\\$\pm$8.900e-03\end{tabular} & \textbf{\begin{tabular}{@{}c@{}}1.012e-01\\$\pm$2.772e-03\end{tabular}} \\
Toxicity & \begin{tabular}{@{}c@{}}3.528e-01\\$\pm$3.861e-03\end{tabular} & \begin{tabular}{@{}c@{}}3.814e-01\\$\pm$2.079e-03\end{tabular} & \begin{tabular}{@{}c@{}}1.407e-01\\$\pm$2.513e-03\end{tabular} & \begin{tabular}{@{}c@{}}1.401e-01\\$\pm$1.007e-03\end{tabular} & \begin{tabular}{@{}c@{}}4.890e-01\\$\pm$2.635e-03\end{tabular} & \textbf{\begin{tabular}{@{}c@{}}1.219e-01\\$\pm$7.051e-03\end{tabular}} \\
SRBCT & \begin{tabular}{@{}c@{}}4.591e-01\\$\pm$4.543e-03\end{tabular} & \begin{tabular}{@{}c@{}}3.854e-01\\$\pm$2.043e-03\end{tabular} & \begin{tabular}{@{}c@{}}5.515e-02\\$\pm$8.084e-03\end{tabular} & \begin{tabular}{@{}c@{}}7.065e-02\\$\pm$3.521e-03\end{tabular} & \begin{tabular}{@{}c@{}}5.216e-01\\$\pm$1.141e-02\end{tabular} & \textbf{\begin{tabular}{@{}c@{}}5.831e-02\\$\pm$3.573e-03\end{tabular}} \\
DLBCL & \begin{tabular}{@{}c@{}}4.332e-01\\$\pm$1.057e-03\end{tabular} & \begin{tabular}{@{}c@{}}4.459e-01\\$\pm$1.861e-02\end{tabular} & \textbf{\begin{tabular}{@{}c@{}}3.717e-05\\$\pm$1.016e-06\end{tabular}} & \begin{tabular}{@{}c@{}}5.332e-02\\$\pm$2.065e-03\end{tabular} & \begin{tabular}{@{}c@{}}4.376e-01\\$\pm$2.325e-03\end{tabular} & \begin{tabular}{@{}c@{}}1.989e-02\\$\pm$4.521e-03\end{tabular} \\
\bottomrule
\end{tabular}
}
\caption{Comparison of IGD values obtained by NSGA-II, MOEA/D, MFFS, MOCDE, CMODE, and MODE-FS on the test datasets.}
\label{tab4}
\end{table*}

In terms of HV, MODE-FS attained superior HV values on $9$ out of 11 datasets, indicating its outstanding ability to construct a well-distributed and diverse Pareto front. A higher HV value represents better objective space coverage and a more balanced trade-off between solution diversity and quality. This improvement is mainly attributed to the diversity enhancement mechanisms integrated into MODE-FS, such as the initialization strategy guided by feature importance and redundancy, dynamic grid partitioning, and iterative subset optimization. In contrast, although the other comparative algorithms exhibited some diversity on certain datasets, they generally exhibited insufficient coverage and poor distribution uniformity. Notably, even on high-dimensional or structurally complex datasets, MODE-FS consistently maintained solution quality, validating its strong global exploration capability and adaptability.

In terms of IGD, MODE-FS achieved the best performance on most datasets, obtaining the best IGD results on 7 datasets. This underscores its strong convergence capability and effectiveness in approximating the true Pareto front. MFFS outperformed MODE-FS on 3 datasets, while NSGA-II achieved the best result on a single dataset. Nevertheless, MODE-FS consistently maintained stable performance across the majority of datasets. Combined with its significant superiority in HV, MODE-FS further consolidated its overall advantage in multiobjective optimization tasks, effectively balancing solution distribution and convergence, and enhancing the discriminative power and adaptability of the selected feature subsets.

In summary, MODE-FS demonstrated consistently superior performance in both HV and IGD compared to the five comparative algorithms. The Pareto fronts generated by MODE-FS were not only widely distributed across the objective space but also closely approximated the true Pareto front, demonstrating strong optimization capabilities in multiobjective feature selection. Particularly when dealing with high-dimensional, redundant, imbalanced, or structurally complex datasets, MODE-FS consistently produced higher-quality and better-structured feature subsets, further verifying its robustness and broad application potential.
\begin{figure*}[!htbp]
    \centering
    \begin{subfigure}{0.3\textwidth}
        \centering
        \includegraphics[width=\linewidth]{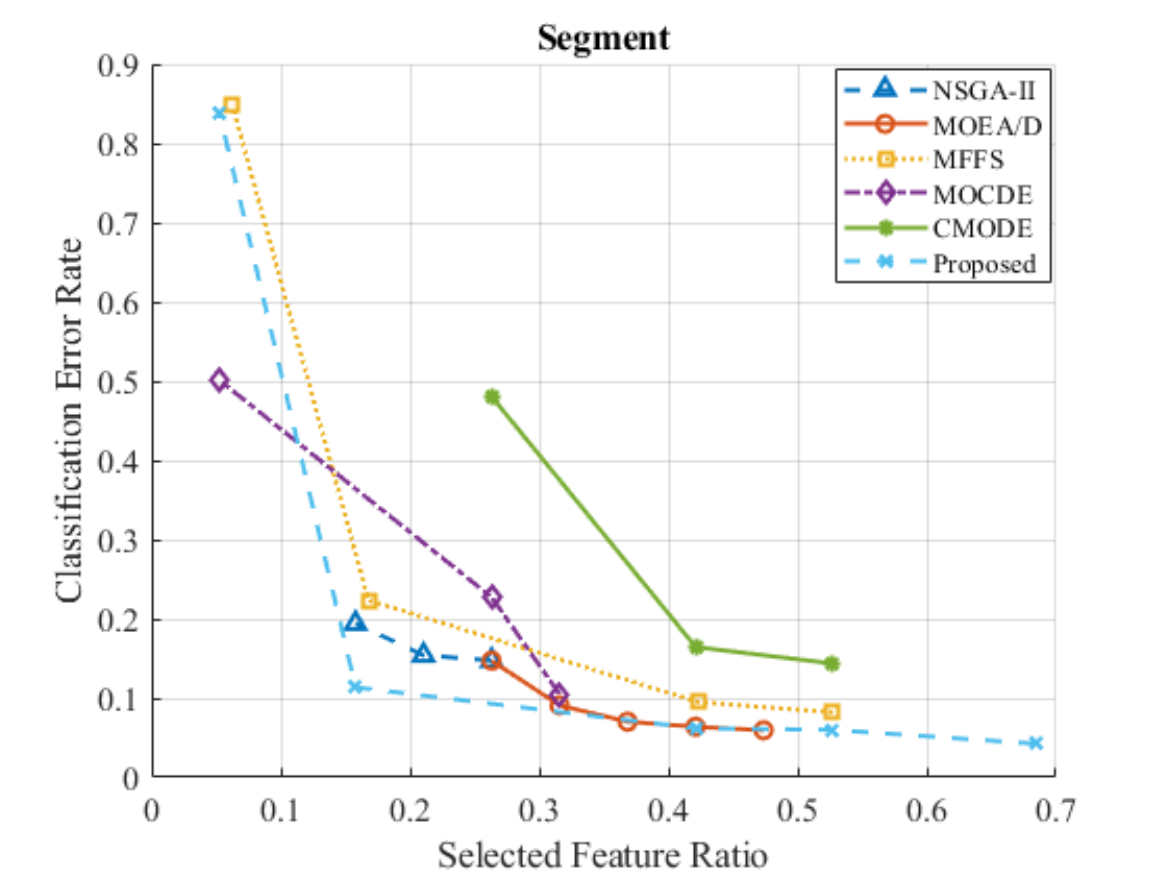}
        \caption{Segment}
    \end{subfigure}
    \hfill
    \begin{subfigure}{0.3\textwidth}
        \centering
        \includegraphics[width=\linewidth]{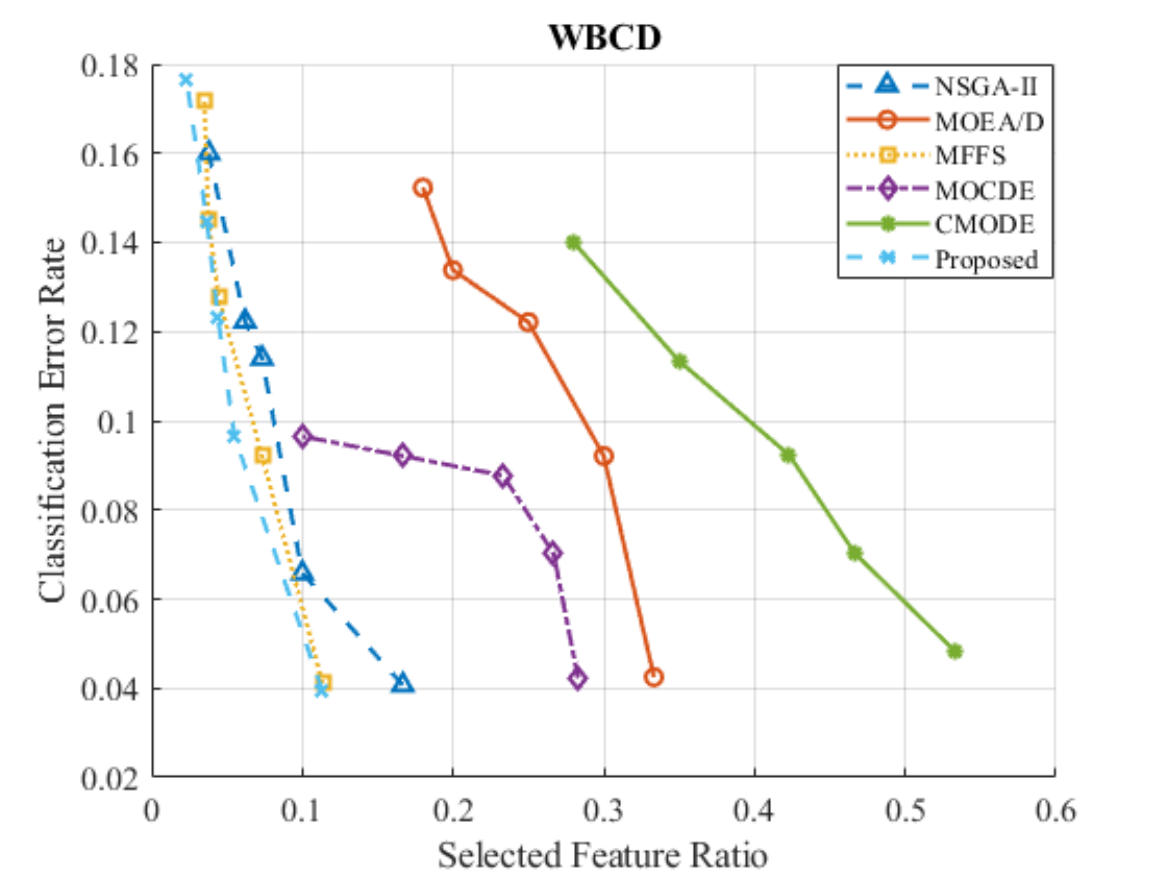}
        \caption{WBCD}
    \end{subfigure}
    \hfill
    \begin{subfigure}{0.3\textwidth}
        \centering
        \includegraphics[width=\linewidth]{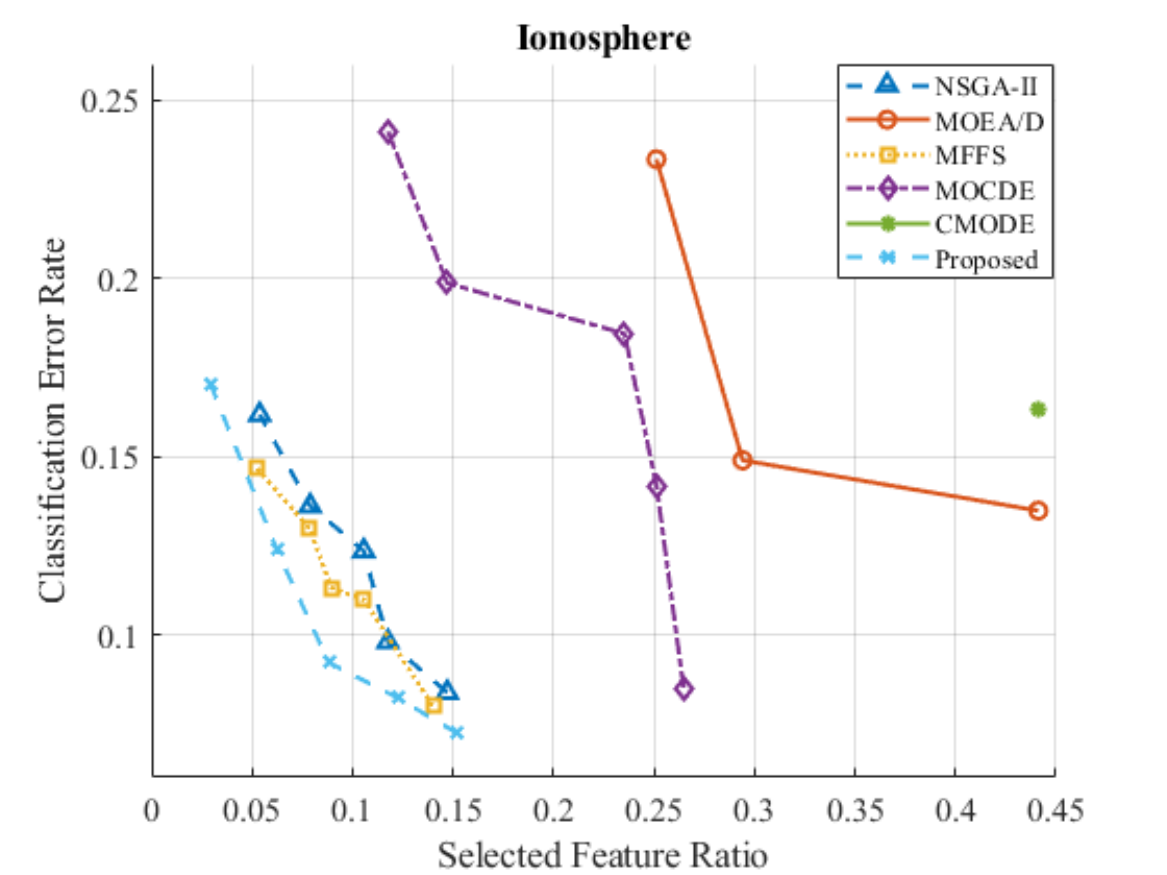}
        \caption{Ionosphere}
    \end{subfigure}

    \vspace{1em}
    \begin{subfigure}{0.3\textwidth}
        \centering
        \includegraphics[width=\linewidth]{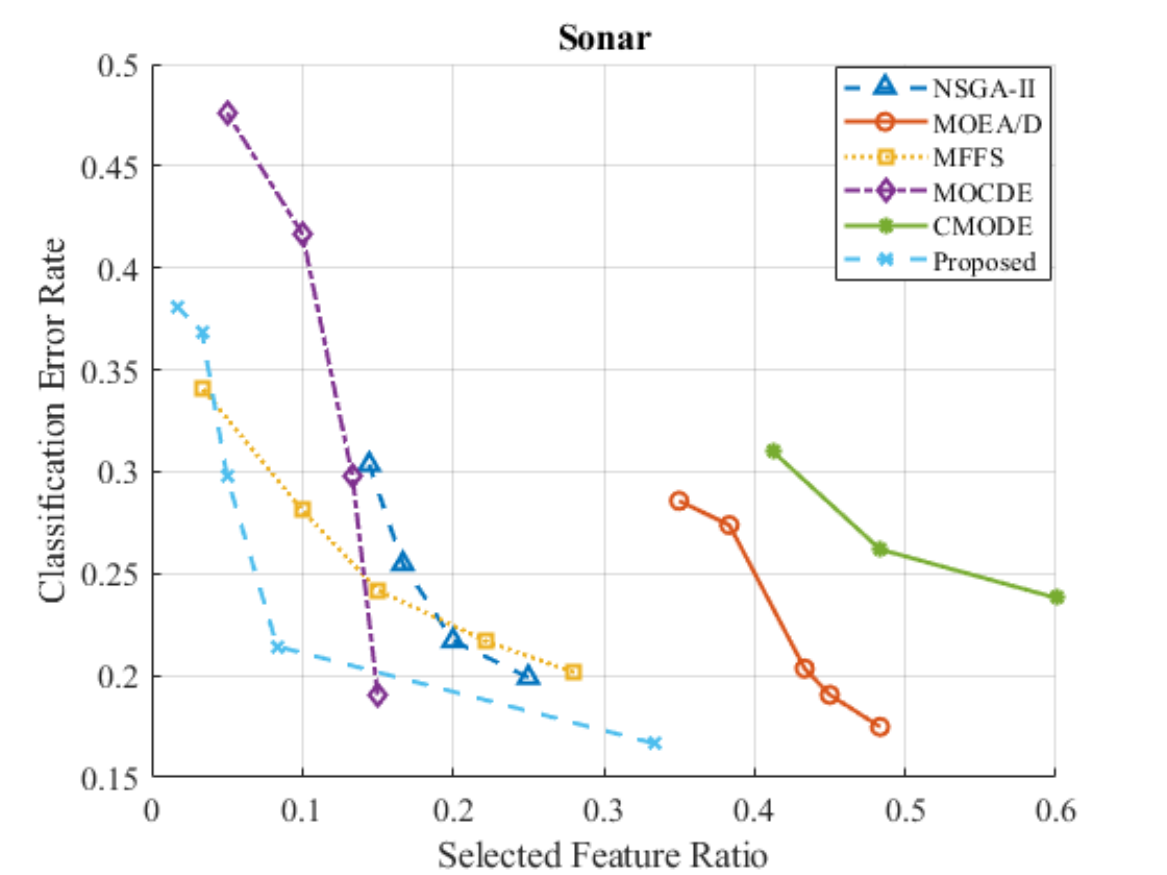}
        \caption{Sonar}
    \end{subfigure}
    \hfill
    \begin{subfigure}{0.3\textwidth}
        \centering
        \includegraphics[width=\linewidth]{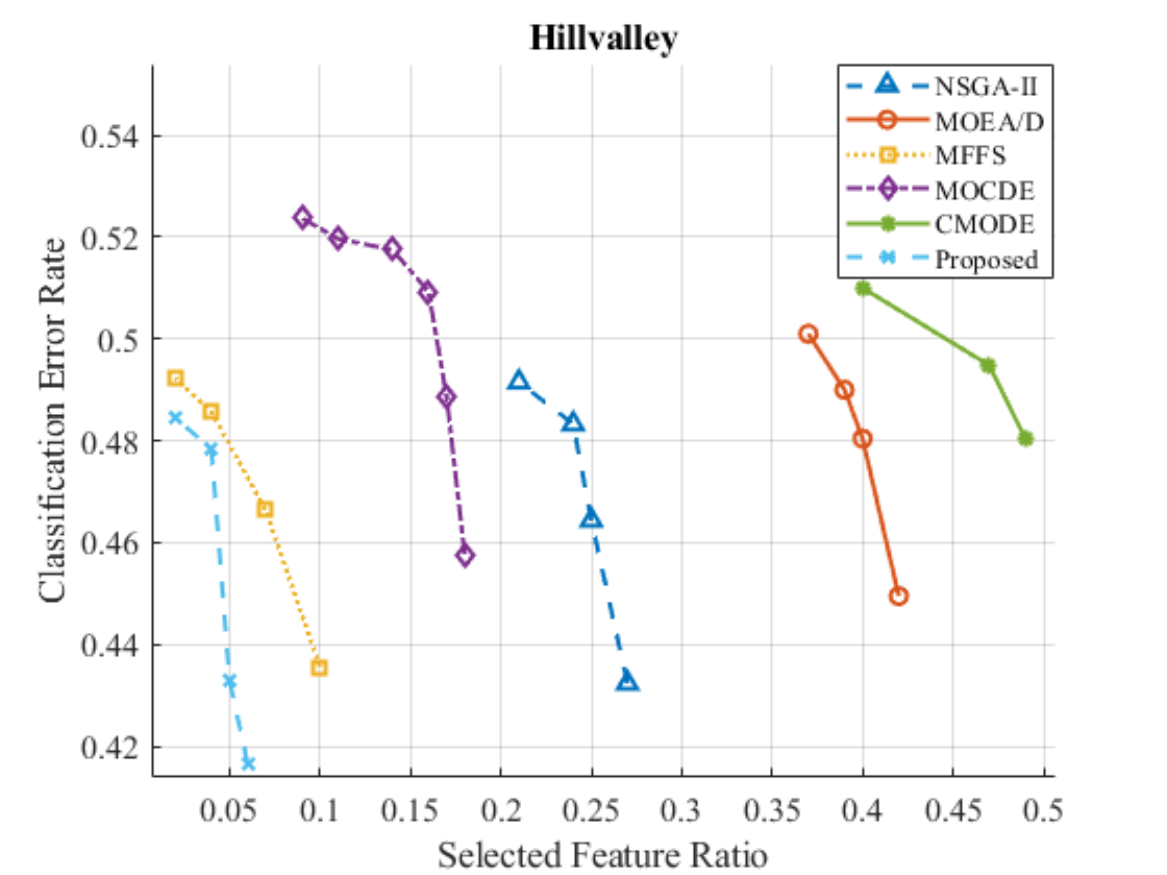}
        \caption{Hillvalley}
    \end{subfigure}
    \hfill
    \begin{subfigure}{0.3\textwidth}
        \centering
        \includegraphics[width=\linewidth]{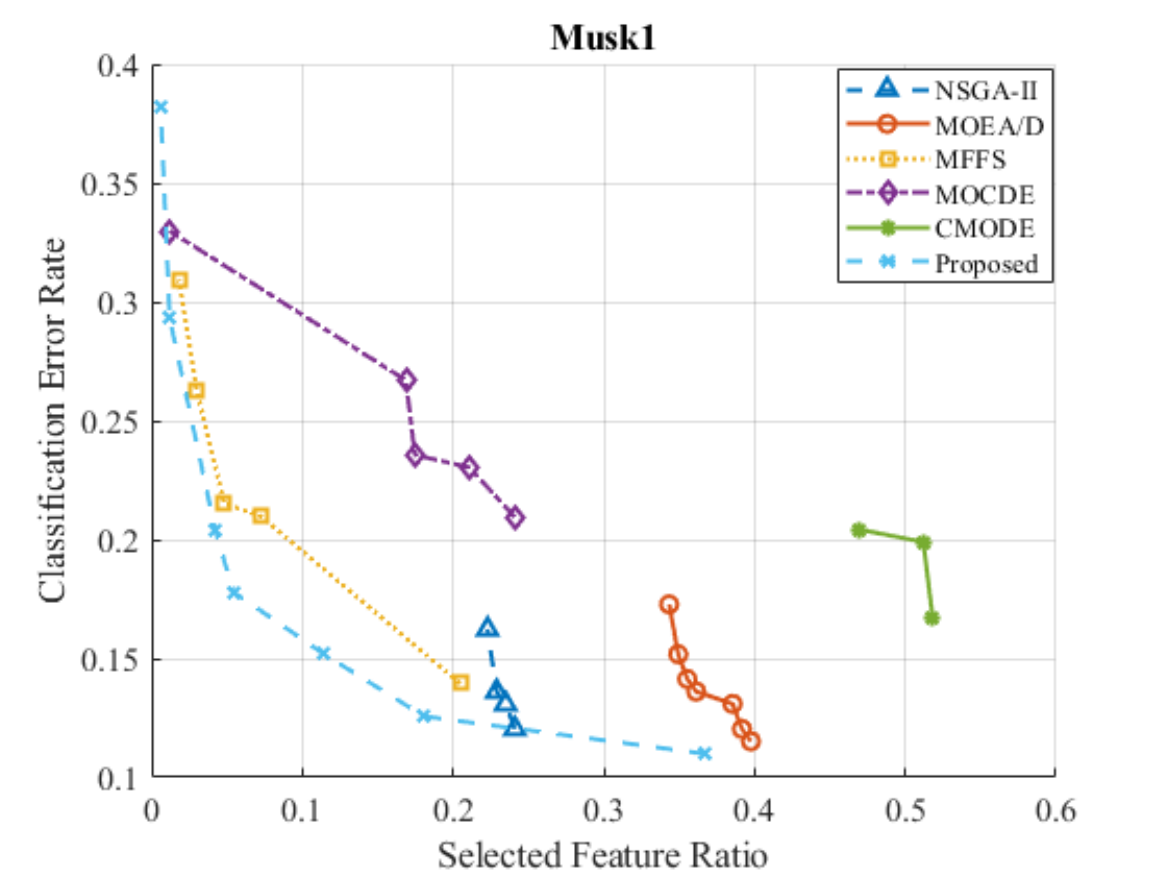}
        \caption{Musk1}
    \end{subfigure}

    \vspace{1em}
    \begin{subfigure}{0.3\textwidth}
        \centering
        \includegraphics[width=\linewidth]{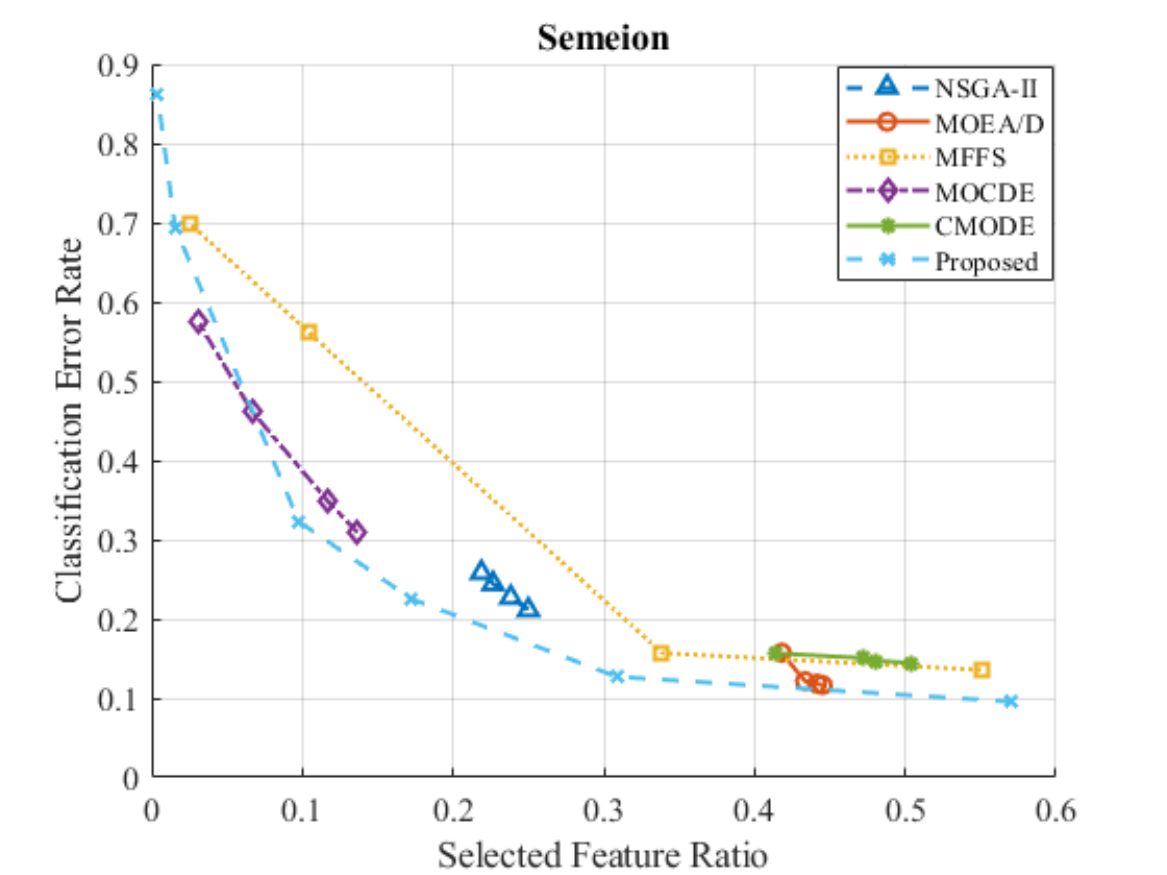}
        \caption{Semeion}
    \end{subfigure}
    \hfill
    \begin{subfigure}{0.3\textwidth}
        \centering
        \includegraphics[width=\linewidth]{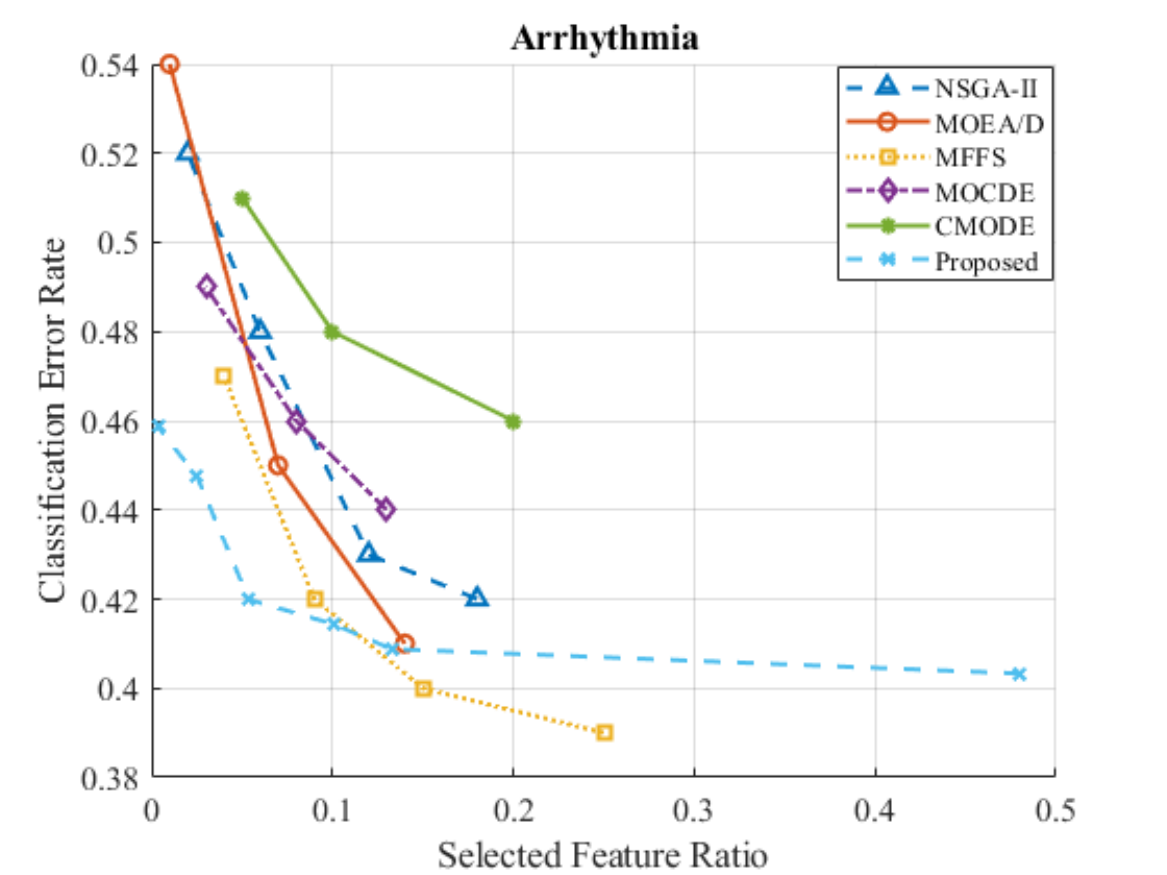}
        \caption{Arrhythmia}
    \end{subfigure}
    \hfill
    \begin{subfigure}{0.3\textwidth}
        \centering
        \includegraphics[width=\linewidth]{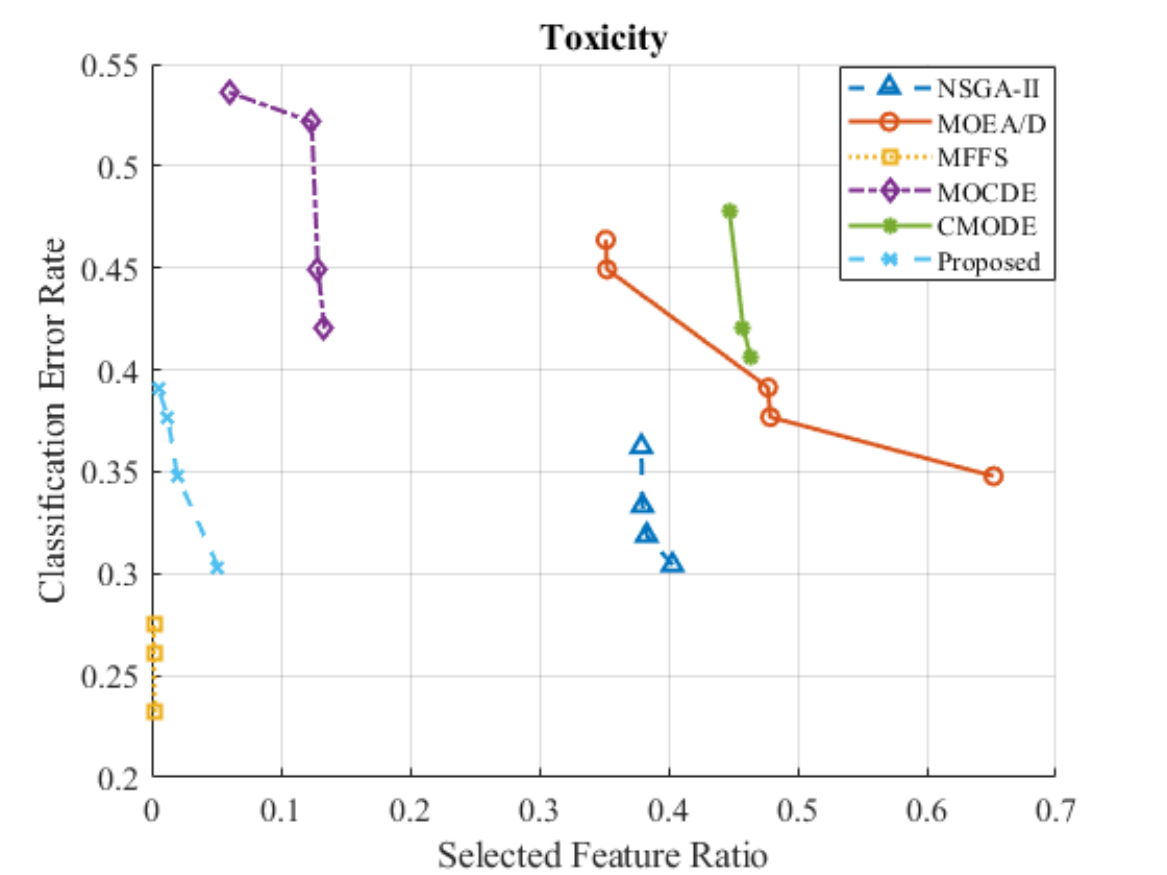}
        \caption{Toxicity}
    \end{subfigure}

    \vspace{1em}
    \begin{subfigure}{0.3\textwidth}
        \centering
        \includegraphics[width=\linewidth]{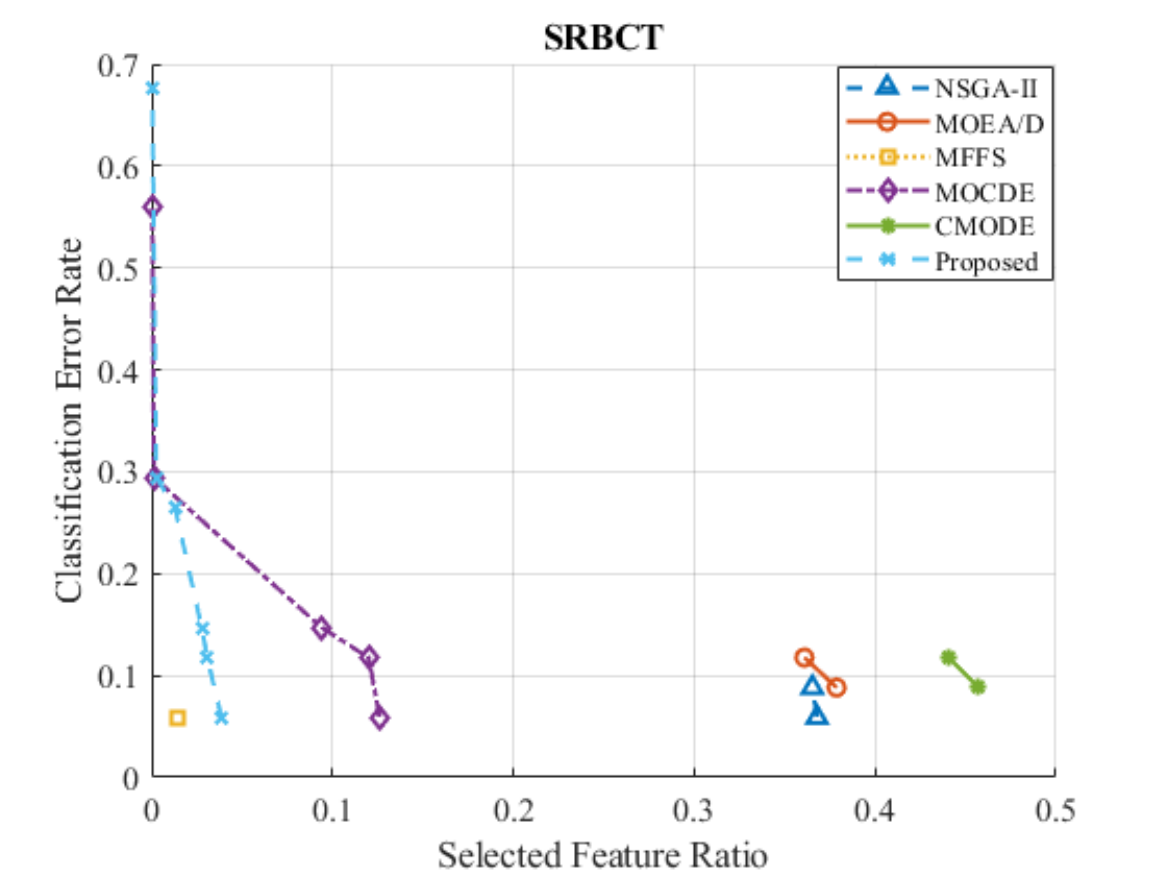}
        \caption{SRBCT}
    \end{subfigure}
    \hspace{0.1\textwidth} 
    \begin{subfigure}{0.3\textwidth}
        \centering
        \includegraphics[width=\linewidth]{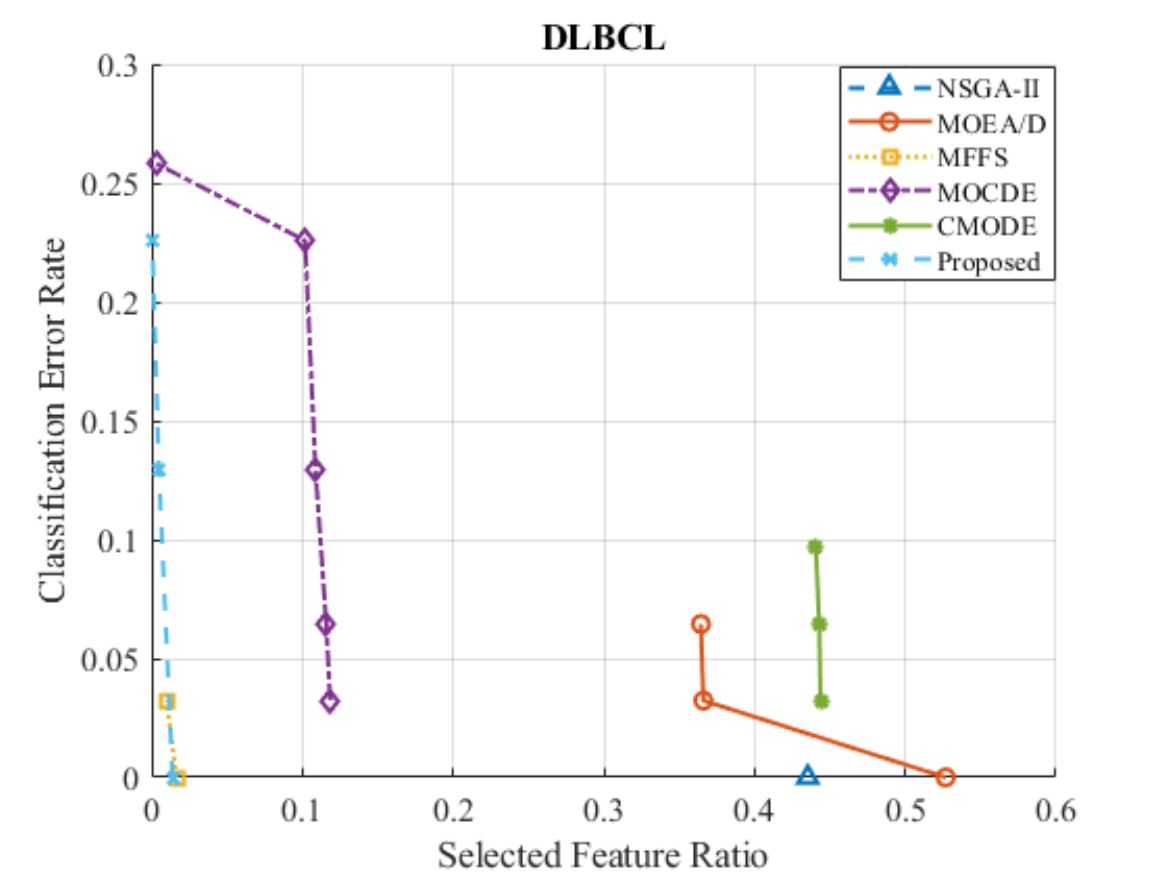}
        \caption{DLBCL}
    \end{subfigure}

    \caption{Distributions of nondominated solutions obtained by each algorithm on test sets in terms of median HV value.}
    \label{fig5}
\end{figure*}

\subsection{Analysis of the Distribution of Nondominated Feature Subsets}
To comprehensively evaluate the performance of MODE-FS in MOFS, Fig.~\ref{fig5} illustrates the distributions of nondominated solutions obtained by MODE-FS and the five comparative algorithms across $11$ representative datasets. The x-axis represents the feature selection ratio, and the y-axis shows the classification error rate. Each solution set represents the Pareto front corresponding to the median HV value obtained from $30$ independent runs of each algorithm.
Overall, MODE-FS demonstrates superior capability in constructing Pareto fronts, achieving a balanced trade-off between feature selection ratio and classification accuracy across most datasets. In particular, MODE-FS consistently achieves superior Pareto fronts compared to other algorithms on eight datasets, including WBCD, Ionosphere, Sonar, Hillvalley, DLBCL, Musk1, Segment, and Semeion. Its nondominated solutions demonstrate excellent performance in both error rate and feature subset size, frequently dominating those produced by other algorithms, thus emphasizing its strong convergence and effective dimensionality reduction ability. For instance, on the WBCD and DLBCL datasets, MODE-FS maintains small feature subsets while achieving exceptionally low error rates, forming continuous and well-structured Pareto fronts. Similarly, on the Musk1, Segment, and Semeion datasets, the nondominated solutions of MODE-FS are densely distributed and broadly cover the Pareto boundary, demonstrating the algorithm’s strong adaptability and stability across diverse feature dimensionalities and sample scales.

MODE-FS exhibits relatively lower performance on the Arrhythmia, Toxicity, and SRBCT datasets. On the Arrhythmia dataset, MODE-FS produces a higher final classification error rate compared to MFFS, but its solution set exhibits a broader distribution in the objective space, thus contributing to greater diversity. On the Toxicity dataset, although MFFS achieves a better overall Pareto front, MODE-FS still provides nondominated solutions that outperform those of other comparative algorithms, demonstrating local advantages. On the SRBCT dataset, MFFS produces only a single nondominated solution point with limited coverage, whereas MODE-FS, although partially dominated, maintains a broader boundary exploration capability. Overall, although MODE-FS solutions are occasionally locally dominated, complete domination is rarely observed. Moreover, MODE-FS consistently produces competitive solution sets with respect to feature selection ratio and classification accuracy, demonstrating stable overall performance and confirming its robustness as a multiobjective feature selection approach.
\section{Conclusion}\label{sec:conclusion}
This study proposes a high-dimensional feature selection algorithm based on multiobjective differential evolution to address key challenges in multiobjective feature selection. The proposed method enhances the diversity of the initial population, improves optimization efficiency, and achieves broader coverage and more uniform distribution of final solutions. A feature correlation matrix and redundancy index are constructed, introducing an FCM-based feature weighting strategy and a cosine similarity-based redundancy assessment mechanism. The initial population is divided into distinct subpopulations, offering more targeted starting points for the search. During the optimization phase, a feature selection mechanism is developed that integrates mutation operations with weight-guided selection, improving solution quality through nondominated sorting and a priority rule based on classification error. Additionally, an adaptive grid refinement strategy is employed to manage dense regions in the objective space, balancing solution diversity and global distribution through individual-level optimization.

Experimental results demonstrate that the proposed approach consistently delivers strong optimization performance across multiple high-dimensional datasets, enhancing solution accuracy while minimizing computational overhead. By fully exploiting structural relationships among features, the method exhibits strong adaptability and generalization capability in intricate situations. Future research will concentrate on dynamic weight adjustment, deeper exploration of inter-feature dependencies, and extending the proposed method to real-time applications, offering novel perspectives and contributions to multiobjective optimization and feature selection research.


\begin{thebibliography}{00}




\bibitem{xue2023_1}
Y. Xue, H. Zhu, F. Neri, A feature selection approach based on NSGA-II with ReliefF, Appl. Soft Comput. 134 (2023) 109987.

\bibitem{hashemi2021_2}
A. Hashemi, M.B. Dowlatshahi, H. Nezamabadi-pour, A pareto-based ensemble of feature selection algorithms, Expert Syst. Appl. 180 (2021) 115130.



\bibitem{jiao2024_3}
R. Jiao, B. Xue, M. Zhang, Learning to Preselection: A Filter-Based Performance Predictor for Multiobjective Feature Selection in Classification, IEEE Trans. Evol. Comput. (2024).

\bibitem{jiao2023_4}
R. Jiao, B. Xue, M. Zhang, Benefiting From Single-Objective Feature Selection to Multiobjective Feature Selection: A Multiform Approach, IEEE Trans. Cybern. 53 (12) (2023) 7773–7786.

\bibitem{wang2023_5}
P. Wang, B. Xue, J. Liang, et al., Feature selection using diversity-based multi-objective binary differential evolution, Inf. Sci. 626 (2023) 586–606.


\bibitem{pan2022_6}
J.S. Pan, N. Liu, S.C. Chu, A competitive mechanism based multi-objective differential evolution algorithm and its application in feature selection, Knowledge-Based Systems 245 (2022) 108582.

\bibitem{deb2002_7}
K. Deb, A. Pratap, S. Agarwal, T. Meyarivan, A fast and elitist multiobjective genetic algorithm: NSGA-II, IEEE Trans. Evol. Comput. 6 (2) (2002) 182–197.

\bibitem{zhang2007_8}
Q. Zhang, H. Li, MOEA/D: A multiobjective evolutionary algorithm based on decomposition, IEEE Trans. Evol. Comput. 11 (6) (2007) 712–731.

\bibitem{tian2017_9}
Y. Tian, R. Cheng, X. Zhang, et al., An indicator-based multiobjective evolutionary algorithm with reference point adaptation for better versatility, IEEE Trans. Evol. Comput. 22 (4) (2017) 609–622.

\bibitem{deb1995_10}
K. Deb, N. Srinivas, Multiobjective optimization using nondominated sorting in genetic algorithms, Evol. Comput. 2 (3) (1995) 221–248.

\bibitem{fister2015_11}
I. Fister, M. Perc, S.M. Kamal, I. Fister, A review of chaos-based firefly algorithms: Perspectives and research challenges, Appl. Math. Comput. 252 (2015) 155–165.

\bibitem{qin2020_12}
S. Qin, C. Sun, G. Zhang, X. He, Y. Tan, A modified particle swarm optimization based on decomposition with different ideal points for many-objective optimization problems, Complex Intell. Syst. 6 (2) (2020) 263–274.

\bibitem{kukkonen2005_13}
S. Kukkonen, J. Lampinen, GDE3: The third evolution step of generalized differential evolution, in: 2005 IEEE Congress on Evolutionary Computation, IEEE CEC 2005. Proceedings, Vol. 1, 2005, pp. 443–450.

\bibitem{wang2023_14}
P. Wang, B. Xue, J. Liang, M. Zhang, Feature clustering-assisted feature selection with differential evolution, Pattern Recognit. 140 (2023) 109523.

\bibitem{hu2023_15}
Y. Hu, M. Lu, X. Li, et al., Differential evolution based on network structure for feature selection, Inf. Sci. 635 (2023) 279–297.


\bibitem{agrawal2023_16}
S. Agrawal, A. Tiwari, B. Yaduvanshi, et al., Feature subset selection using multimodal multiobjective differential evolution, Knowledge-Based Systems 265 (2023) 110361.

\bibitem{yu2025_17}
F. Yu, J. Guan, H. Wu, et al., Multi-population differential evolution approach for feature selection with mutual information ranking, Expert Syst. Appl. 260 (2025) 125404.

\bibitem{yu2024_18}
X. Yu, Z. Hu, W. Luo, et al., Reinforcement learning-based multi-objective differential evolution algorithm for feature selection, Inf. Sci. 661 (2024) 120185.

\bibitem{yue2018_19}
C. Yue, B. Qu, J. Liang, A multiobjective particle swarm optimizer using ring topology for solving multimodal multiobjective problems, IEEE Trans. Evol. Comput. 22 (5) (2018) 805–817.

\bibitem{kosko1986_20}
B. Kosko, Fuzzy cognitive maps, Int. J. Man-Mach. Stud. 24 (1) (1986) 65–75.

\bibitem{xue2014_21}
B. Xue, W. Fu, M. Zhang, Differential evolution for multi-objective feature selection in classification, in: Proc. Annu. Conf. Genet. Evol. Comput., 2014, pp. 83–84.

\bibitem{zhang2015_22}
Y. Zhang, D. Gong, M. Rong, Multi-objective differential evolution algorithm for multi-label feature selection in classification, in: Proc. Int. Conf. Swarm Intell., 2015, pp. 339–345.

\bibitem{salesi2021_23}
S. Salesi, G. Cosma, M. Mavrovouniotis, Taga: Tabu asexual genetic algorithm embedded in a filter/filter feature selection approach for high-dimensional data, Inf. Sci. 565 (2021) 105–127.

\bibitem{li2022_24}
T. Li, Z. Zhan, J. Xu, Q. Yang, Y. Ma, A binary individual search strategy-based bi-objective evolutionary algorithm for high-dimensional feature selection, Inf. Sci. 610 (2022) 651–673.

\bibitem{xu2020_25}
H. Xu, B. Xue, M. Zhang, A duplication analysis based evolutionary algorithm for bi-objective feature selection, IEEE Trans. Evol. Comput. 25 (2) (2020) 205–218.

\bibitem{wang2021cyb_26}
P. Wang, B. Xue, J. Liang, et al., Multiobjective differential evolution for feature selection in classification, IEEE Trans. Cybern. 53 (7) (2021) 4579–4593.

\bibitem{wang2022niching_27}
P. Wang, B. Xue, J. Liang, et al., Differential evolution-based feature selection: A niching-based multiobjective approach, IEEE Trans. Evol. Comput. 27 (2) (2022) 296–310.

\bibitem{ahadzadeh2023_28}
B. Ahadzadeh, M. Abdar, F. Safara, et al., SFE: a simple, fast, and efficient feature selection algorithm for high-dimensional data, IEEE Trans. Evol. Comput. 27 (6) (2023) 1896–1911.

\bibitem{song2021_29}
X.F. Song, Y. Zhang, D.W. Gong, et al., A fast hybrid feature selection based on correlation-guided clustering and particle swarm optimization for high-dimensional data, IEEE Trans. Cybern. 52 (9) (2021) 9573–9586.

\bibitem{cheng2021_30}
F. Cheng, F. Chu, Y. Xu, et al., A steering-matrix-based multiobjective evolutionary algorithm for high-dimensional feature selection, IEEE Trans. Cybern. 52 (9) (2021) 9695–9708.

\bibitem{kwak2021_31}
B.I. Kwak, M.L. Han, H.K. Kim, Cosine similarity based anomaly detection methodology for the CAN bus, Expert Syst. Appl. 166 (2021) 114066.

\bibitem{dua2017_32}
D. Dua, C. Graff, UCI Machine Learning Repository, 2017. [Online]. Available: http://archive.ics.uci.edu/ml

\bibitem{tran2019_33}
B. Tran, B. Xue, M. Zhang, Variable-length particle swarm optimization for feature selection on high-dimensional classification, IEEE Trans. Evol. Comput. 23 (3) (2019) 473–487.

\bibitem{zitzler1999_34}
E. Zitzler, Evolutionary algorithms for multiobjective optimization: Methods and applications, Ph.D. dissertation, Dept. Comput. Eng. Netw. Lab., Swiss Federal Inst. Technol., Zürich, Switzerland, 1999.

\bibitem{xu2020gecco_35}
H. Xu, B. Xue, M. Zhang, Segmented initialization and off-spring modification in evolutionary algorithms for bi-objective feature selection, in: Proc. Genet. Evol. Comput. Conf. (GECCO), 2020, pp. 444–452.

\bibitem{bosman2003_36}
P.A.N. Bosman, D. Thierens, The balance between proximity and diversity in multiobjective evolutionary algorithms, IEEE Trans. Evol. Comput. 7 (2) (2003) 174–188.


\end{thebibliography}
\end{document}